%% file: main.tex
\documentclass{article}

\usepackage{arxiv}

\usepackage[utf8]{inputenc} 
\usepackage[T1]{fontenc}    
\usepackage{hyperref}       
\usepackage{url}            
\usepackage{booktabs}       
\usepackage{amsfonts}       
\usepackage{nicefrac}       
\usepackage{microtype}      
\usepackage{graphicx}
\usepackage[
    backend=biber,
    bibstyle=ieee,
    citestyle=numeric-comp,
    hyperref=true,
    date=year,
    bibencoding=inputenc,
    uniquename=init,
    giveninits=true
]{biblatex}
\usepackage{doi}

\usepackage{longtable}
\usepackage{amsmath}
\usepackage{siunitx}
\usepackage{adjustbox}
\usepackage{subcaption}
\usepackage{import}
\usepackage{breakcites}
\usepackage[super]{nth}
\usepackage{bm}
\usepackage[disable]{todonotes}

\makeatletter
\newcommand\footnoteref[1]{\protected@xdef\@thefnmark{\ref{#1}}\@footnotemark}
\makeatother

\title{Review of automated time series forecasting pipelines}

\author{
    \href{https://orcid.org/0000-0002-9320-5341}{\includegraphics[scale=0.06]{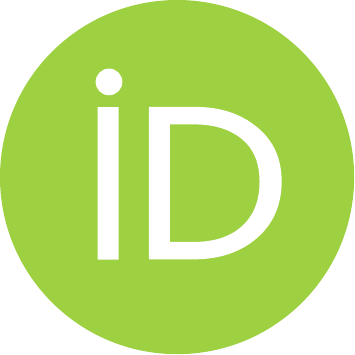}\hspace{1mm}
    Stefan Meisenbacher}\\
	Institute for Automation and Applied Informatics\\
	Karlsruhe Institute of Technology\\
	Eggenstein-Leopoldshafen, 76344, Germany\\
	\texttt{stefan.meisenbacher@kit.edu}\\
	\And
    \href{https://orcid.org/0000-0002-3776-2215}{\includegraphics[scale=0.06]{orcid.pdf}\hspace{1mm}
    Marian Turowski}\\
	Institute for Automation and Applied Informatics\\
	Karlsruhe Institute of Technology\\
	Eggenstein-Leopoldshafen, 76344, Germany\\
	\And
    \href{https://orcid.org/0000-0002-9197-1739}{\includegraphics[scale=0.06]{orcid.pdf}\hspace{1mm}
    Kaleb Phipps}\\
	Institute for Automation and Applied Informatics\\
	Karlsruhe Institute of Technology\\
	Eggenstein-Leopoldshafen, 76344, Germany\\
	\And
    \href{https://orcid.org/0000-0002-3573-2872}{\includegraphics[scale=0.06]{orcid.pdf}\hspace{1mm}
    Martin Rätz}\\
	Institute for Energy Efficient Buildings and Indoor Climate\\
	RWTH Aachen University\\
	Aachen, 52062, Germany\\
	\And
    \href{https://orcid.org/0000-0002-6106-6607}{\includegraphics[scale=0.06]{orcid.pdf}\hspace{1mm}
    Dirk Müller}\\
	Institute for Energy Efficient Buildings and Indoor Climate\\
	RWTH Aachen University\\
	Aachen, 52062, Germany\\
	Forschungszentrum Jülich GmbH\\
	Institute of Energy and Climate Research\\
	Energy Systems Engineering (IEK-10)\\
	Jülich, 52425, Germany
	\And
    \href{https://orcid.org/0000-0002-3572-9083}{\includegraphics[scale=0.06]{orcid.pdf}\hspace{1mm}
    Veit Hagenmeyer}\\
	Institute for Automation and Applied Informatics\\
	Karlsruhe Institute of Technology\\
	Eggenstein-Leopoldshafen, 76344, Germany\\
	\And
    \href{https://orcid.org/0000-0001-9100-5496}{\includegraphics[scale=0.06]{orcid.pdf}\hspace{1mm}
    Ralf Mikut}\\
	Institute for Automation and Applied Informatics\\
	Karlsruhe Institute of Technology\\
	Eggenstein-Leopoldshafen, 76344, Germany\\
}

\date{}

\renewcommand{\headeright}{Meisenbacher \textit{et al.}}
\renewcommand{\undertitle}{}
\renewcommand{\shorttitle}{Review of automated time series forecasting pipelines}

\hypersetup{
    pdftitle={Review of automated time series forecasting pipelines},
    pdfsubject={automated time series forecasting pipelines},
    pdfauthor={Meisenbacher et al.},
    pdfkeywords={
        Automated Machine Learning,
        Time Series Forecasting,
        AutoML,
        Pipeline,
        Pre-processing,
        Feature Engineering,
        Hyperparameter Optimization,
        Forecasting Method Selection,
        Ensemble
    },
    hidelinks
}

\bibliography{references}

\begin{document}
\maketitle
\nocite{*}  
\begin{abstract}
    Time series forecasting is fundamental for various use cases in different domains such as energy systems and economics.
    Creating a forecasting model for a specific use case requires an iterative and complex design process.
    The typical design process includes the five sections (1) data pre-processing, (2) feature engineering, (3) hyperparameter optimization, (4) forecasting method selection, and (5) forecast ensembling, which are commonly organized in a pipeline structure.
    One promising approach to handle the ever-growing demand for time series forecasts is automating this design process.
    The present paper, thus, analyzes the existing literature on automated time series forecasting pipelines to investigate how to automate the design process of forecasting models.
    Thereby, we consider both Automated Machine Learning (AutoML) and automated statistical forecasting methods in a single forecasting pipeline.
    For this purpose, we firstly present and compare the proposed automation methods for each pipeline section.
    Secondly, we analyze the automation methods regarding their interaction, combination, and coverage of the five pipeline sections.
    For both, we discuss the literature, identify problems, give recommendations, and suggest future research.
    This review reveals that the majority of papers only cover two or three of the five pipeline sections.
    We conclude that future research has to holistically consider the automation of the forecasting pipeline to enable the large-scale application of time series forecasting.
\end{abstract}

\keywords{
    Automated Machine Learning \and
    Time Series Forecasting \and
    AutoML \and
    Pipeline \and
    Pre-processing \and
    Feature Engineering \and
    Hyperparameter Optimization \and
    Forecasting Method Selection \and
    Ensemble
}

\section{Introduction}
\label{sec:Introduction}
\input{content/01-introduction}


\section{Methodology}
\label{sec:Methodology}
\input{content/03-methodology}

\section{Time Series Forecasting}
\label{sec:TimeSeriesForecasting}
\input{content/04-time-series-forecasting}

\section{Data Pre-processing}
\label{sec:PreProcessing}
\input{content/05-pre-processing}

\section{Feature Engineering}
\label{sec:FeatureEngineering}
\input{content/06-feature-engineering}

\section{Hyperparameter Optimization}
\label{sec:HyperparameterOptimization}
\input{content/07-hyperparameter-optimization}

\section{Forecasting Method Selection and Ensembling}
\label{sec:SelectionEnsembling}
\input{content/08-selection-ensembling}

\section{Automated Forecasting Pipeline}
\label{sec:Pipeline}
\input{content/09-forecasting-pipeline}

\section{Conclusions}
\label{sec:Conclusions}
\input{content/10-conclusions}

\section*{Acknowledgements}

Conceptualization and methodology:                      S. M., M. T., K. P., M. R., R. M.;
Literature search, screening, analysis, and review:     S. M.;
Review criteria and table design:                       S. M., M. T., K. P., M. R.;
Structure:                                              S. M., M. T., K. P.;
Visualization and layout:                               S. M.;
Writing -- original draft preparation:                  S. M.; 
Writing -- review and editing:                          S. M., M. T., K. P., M. R., D. M., V. H, R. M.;
Supervision and funding acquisition:                    D. M., V. H., R. M.;
All authors have read and agreed to the published version of the article.

The Helmholtz Association’s Initiative and Networking Fund through Helmholtz AI; the Helmholtz Association under the Program “Energy System Design”; the German Research Foundation (DFG) as part of the Research Training Group 2153 “Energy Status Data: Informatics Methods for its Collection, Analysis and Exploitation”; Federal Ministry for Economic Affairs and Energy (BMWi), promotional reference 03ET1568.

\section*{Conflict of Interest}

The authors declare no conflict of interest.

\section*{Supporting Information}

Search-term for title, abstract, and keywords:
\begin{description}
    \item (
    \begin{description}
        \item (
        \begin{description}
            \item automl OR
            \item ``automat* model*'' OR
            \item ``automat* deep learning'' OR
            \item ``automat* machine learning'' OR
            \item ``automat* forecast*'' OR
            \item ``automat* predict*''
        \end{description}
        \item )
        \item AND (``time-series'')
    \end{description}
    \item )
    \item OR ``automat* time-series''
\end{description}

The search-term based exploration was conducted on 4/29/2021 using \textit{Scopus} and yielded 359 papers.
The forward search was conducted on 6/9/2021 and yielded 1572 additional papers.
The backward search was conducted on 8/17/2021 and yielded 580 additional papers.

\section*{Acronyms}

\begin{longtable}{ p{.15\linewidth}  p{.85\linewidth} } 
ACF&        AutoCorrelation Function\\
ADF&        Augmented Dickey-Fuller\\
AIC&        Akaike Information Criterion\\
ANN&        Artificial Neural Network\\
AR&         AutoRegression\\
ARMA&       AutoRegressive Moving Average\\
ARIMA&      AutoRegressive Integrated Moving Average\\
AutoML&		Automated Machine Learning\\
BE&         Backward Elimination\\
BIC&        Bayesian Information Criterion\\
BO&			Bayesian Optimization\\
BSTS&       Bayesian Structural Time Series\\
CASH&		Combined Algorithm Selection and Hyperparameter Optimization\\
CNN&        Convolutional Neural Network\\
DBN&        Deep Belief Network\\
DBSCAN&     Density-Based Spatial Clustering of Applications with Noise\\
DEA&        Differential Evolution Algorithm\\
DES&        Double Exponential Smoothing\\
DT&         Decision Tree\\
E2E&        End-to-End\\
EA&         Evolutionary Algorithm\\
EDA&        Estimation Distribution Algorithm\\
ELM&        Extreme Learning Machine\\
ENN&        Elman Neural Network\\
ES&         Exponential Smoothing\\
ETS&        Error Trend Seasonality\\
FCNN&       Fully Connected Neural Network\\
FNN&        Fuzzy Neural Network\\
FS&         Forward Selection\\
GA&         Genetic Algorithm\\
GBM&        Gradient Boosting Machine\\
GOF&        Goodness-Of-Fit\\
GP&			Gaussian Process\\
GRNN&       General Regression Neural Network\\
HPO&		HyperParameter Optimization\\
IC&         Information Criteria\\
IMA&        Integrated Moving Average\\
INF&        Iterative Neural Filter\\
IQR&        Inter-Quartile Range\\
kNN&        k-Nearest Neighbors\\
KPSS&       Kwiatkowski--Phillips--Schmidt--Shin\\
KS&         Kolmogorov--Smirnov\\
LASSO&      Least Absolute Shrinkage and Selection Operator\\
LC&         Locally Constant\\
LDA&        Linear Discriminant Analysis\\
LL&         Locally Linear\\
LogR&       Logistic Regression\\
LR&         Linear Regression\\
LSTM&       Long Short-Term Memory\\
MA&         Moving Average\\
MAD&        Median Absolute Deviation\\
MAE&        Mean Average Error\\
MAPE&       Mean Absolute Percentage Error\\
MARS&       Multivariate Adaptive Regression Splines\\
MASE&       Mean Absolute Scaled Error\\
MIQP&       Mixed Integer Quadratic Programming\\
MKL&        Multiple Kernel Learning\\
MLP&        MultiLayer Perceptron\\
MSE&        Mean Squared Error\\
NLP&        Non-Linear Programming\\
NNetAR&     Neural Network AutoRegression\\
OSCB&       Osborn--Chui--Smith--Birchenhall\\
PACF&       Partial Auto-Correlation Function\\
PCA&        Principal Component Analysis\\
PR&         Polynomial Regression\\
PSO&        Particle Swarm Optimization\\
RBFNN&      Radial Basis Function Neural Network\\
RF&         Random Forest\\
RMSE&       Root Mean Squared Error\\
RNN&        Recurrent Neural Network\\
RPaRT&      Recursive Partitioning and Regression Trees\\
RW&         Random Walk\\
sARIMA&     seasonal AutoRegressive Integrated Moving Average\\
SES&        Simple Exponential Smoothing\\
SETARMA&    Self Exciting Threshold AutoRegressive Moving Average\\
sMAPE&      symmetric Mean Absolute Percentage Error\\
STL&        Seasonal and Trend decomposition using Loess\\
SVD&        Singular Value Decomposition\\
SVM&        Support Vector Machine\\
SVR&        Support Vector Regression\\
TBATS&      Trigonometric Box-Cox transform, ARMA errors \& Trend and Seasonal components\\
TES&        Triple Exponential Smoothing\\
TPE&		Tree Parzen Estimator\\
UC&         Unexplored Component\\
VAR&        Vector AutoRegression
\end{longtable}

\newpage

\section*{Implementations of Forecasting Methods}

\begin{longtable}{ p{.3\linewidth}  p{.7\linewidth} } 
autoARIMA&
    [\href{https://pkg.robjhyndman.com/forecast/reference/auto.arima.html}{R}]
    [\href{https://www.sktime.org/en/stable/api_reference/auto_generated/sktime.forecasting.arima.AutoARIMA.html}{Python}]
\\
autoTheta&
    [\href{https://github.com/vangspiliot/AutoTheta}{R}]
\\
AR, MA, ARMA, (s)ARIMA&
    [\href{https://pkg.robjhyndman.com/forecast/reference/Arima.html}{R}]
    [\href{https://www.sktime.org/en/stable/api_reference/auto_generated/sktime.forecasting.arima.ARIMA.html}{Python}]
\\
BSTS&
    [\href{https://www.rdocumentation.org/packages/bsts/versions/0.9.7/topics/bsts}{R}]
\\
ELM&
    [\href{https://www.rdocumentation.org/packages/nnfor/versions/0.9.6/topics/forecast.elm}{R}]
\\
ETS&
    [\href{https://pkg.robjhyndman.com/forecast/reference/ets.html}{R}]
    [\href{https://www.sktime.org/en/stable/api_reference/auto_generated/sktime.forecasting.ets.AutoETS.html}{Python}]
\\
GRNN&
    [\href{https://cran.r-project.org/web/packages/tsfgrnn/vignettes/tsfgrnn.html}{R}]
\\
kNN&
    [\href{https://cran.r-project.org/web/packages/tsfknn/vignettes/tsfknn.html}{R}]
\\
MLP INF&
    [\href{https://www.rdocumentation.org/packages/nnfor/versions/0.9.6/topics/forecast.mlp}{R}]
\\
NNetAR&
    [\href{https://pkg.robjhyndman.com/forecast/reference/nnetar.html}{R}]
\\
Prophet&
    [\href{https://facebook.github.io/prophet/docs/quick_start.html#r-api}{R}]
    [\href{https://facebook.github.io/prophet/docs/quick_start.html#python-api}{Python}]
\\
RPaRT&
    [\href{https://www.rdocumentation.org/packages/rpart/versions/4.1-15/topics/rpart}{R}]
\\
RW, dRW, sRW&
    [\href{https://pkg.robjhyndman.com/forecast/reference/naive.html}{R}]
    [\href{https://www.sktime.org/en/stable/api_reference/auto_generated/sktime.forecasting.naive.NaiveForecaster.html}{Python}]
\\
SES, DES, TES&
    [\href{https://pkg.robjhyndman.com/forecast/reference/ses.html}{R}]
    [\href{https://www.sktime.org/en/stable/api_reference/auto_generated/sktime.forecasting.exp_smoothing.ExponentialSmoothing.html}{Python}]
\\
TBATS&
    [\href{https://pkg.robjhyndman.com/forecast/reference/tbats.html}{R}]
    [\href{https://www.sktime.org/en/stable/api_reference/auto_generated/sktime.forecasting.tbats.TBATS.html}{Python}]
\\
Telescope&
    [\href{https://github.com/DescartesResearch/telescope}{R}]
\\
Theta&
    [\href{https://pkg.robjhyndman.com/forecast/reference/thetaf.html}{R}]
    [\href{https://www.sktime.org/en/stable/api_reference/auto_generated/sktime.forecasting.theta.ThetaForecaster.html}{Python}]
\\
VAR&
    [\href{https://www.rdocumentation.org/packages/vars/versions/1.5-6/topics/VAR}{R}]
    [\href{https://www.statsmodels.org/dev/vector_ar.html}{Python}]
\end{longtable}

\printbibliography

\end{document}

%% file: content/01-introduction.tex
One of the most prominent forms of collected data are time series.
In a time series, the data is arranged sequentially, and each value is explicitly time-stamped, with information such as date and time (i.\,e, value and time stamp constitute an observation).
The progression of a time series over a certain period of time in the future, also known as forecast horizon, is the subject of time series forecasting.
Time series forecasting is applied with various forecast horizons at different temporal scales and aggregation levels in various domains~\cite{Hyndman2021}.
Exemplary use cases from different domains are the following:
    sales forecasting for inventory optimization (\textit{just-in-time supply chain}, \cite{Boone2019}),
    forecasting the generation of renewable energy and the electricity demand in an area to balance the power grid load (\textit{smart grids}, \cite{Ahmad2020}), and
    forecasting the spread of the novel coronavirus COVID-19 (\textit{pandemic control}, \cite{Rahimi2021}).
As the number and importance of use cases grow, the demand for time series forecasts is increasing steadily.

Designing a time series forecast for a particular use case typically incorporates five sections.
The first section of the design process is the data pre-processing to transform the raw data into a desirable form for the forecasting method~\cite{Wang2020, Shaukat2021}.
The second section is the feature engineering, which aims to extract hidden characteristics of the considered time series or to identify useful exogenous information for the forecasting method~\cite{Zebari2020}.
Each forecasting method contains hyperparameters that have to be set by the data scientist. 
Therefore, the third section, the HyperParameter Optimization (HPO), intends to improve the forecast accuracy over the default hyperparameter configuration~\cite{Hutter2019}.
Apart from the HPO, selecting the most suitable forecasting method is crucial for the forecast accuracy and is addressed in the fourth section~\cite{Zoeller2021}.
The fifth section aims to increase the robustness of the forecast by forecast ensembling~\cite{Hajirahimi2019}, i.\,e., bundling multiple forecasts of different forecasting models to avoid occasional poor forecasts~\cite{Shaub2020}.

\begin{figure}[h]
	\centering
	\includegraphics[width=\linewidth]{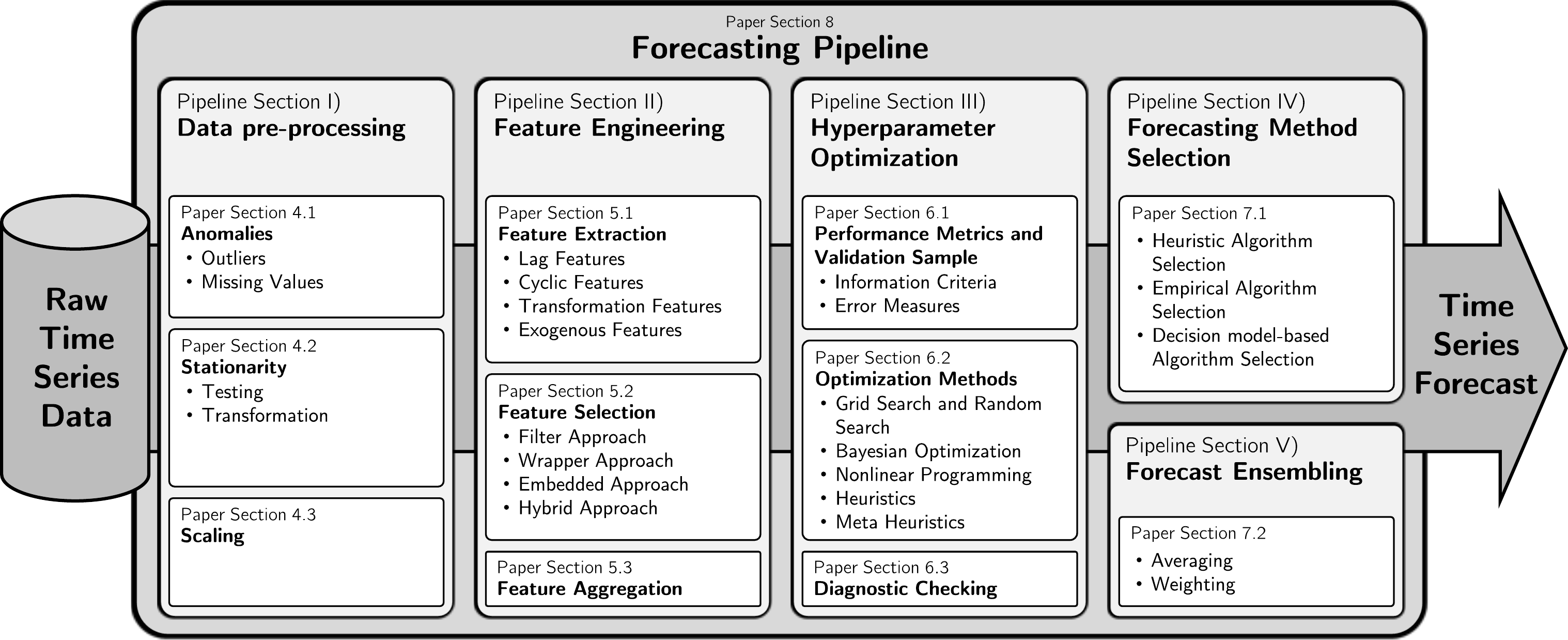}
	\caption{The forecasting pipeline systematizes the design process for time series forecasting using five pipeline sections.}
	\label{fig:TimeSeriesPipeline}
\end{figure}

The above sections of the design process are commonly organized in a pipeline structure as shown in~\autoref{fig:TimeSeriesPipeline}.
Manually tailoring the forecasting pipeline to a specific use case is time-consuming and challenging because selecting appropriate methods for the pipeline sections is iterative and requires expert knowledge.
This expert knowledge is particularly crucial, as the forecast accuracy is sensitive to various design decisions~\cite{Hutter2019}.
It is also foreseeable that the number of knowledgeable data scientists cannot handle the ever-growing demand for time series forecasts in the future.
Therefore, increasing the efficiency of the design process by automation is required~\cite{Tuggener2019}.

To automate design decisions and remove the data scientist from the iterative design process, a variety of automation methods are available for each section of the forecasting pipeline.\footnote{
    In addition to automating the design process, automating the operation of forecasts is proposed by \cite{Meisenbacher2021}, which includes self-monitoring and automatic model adaption as forecast accuracy decreases.
}
The sequential organization of these automation methods and the management of the data flow can be realized by creating a pipeline \cite{pyWATTS, Sktime}.
Running the created forecasting pipeline trains a forecasting method -- e.\,g. a Linear Regression (LR) -- with historical time series data and results in a parameterized forecasting model.
Thereby, the pipeline automates the design process.

\begin{figure}[hb!]
	\centering
	\includegraphics[width=11.6cm]{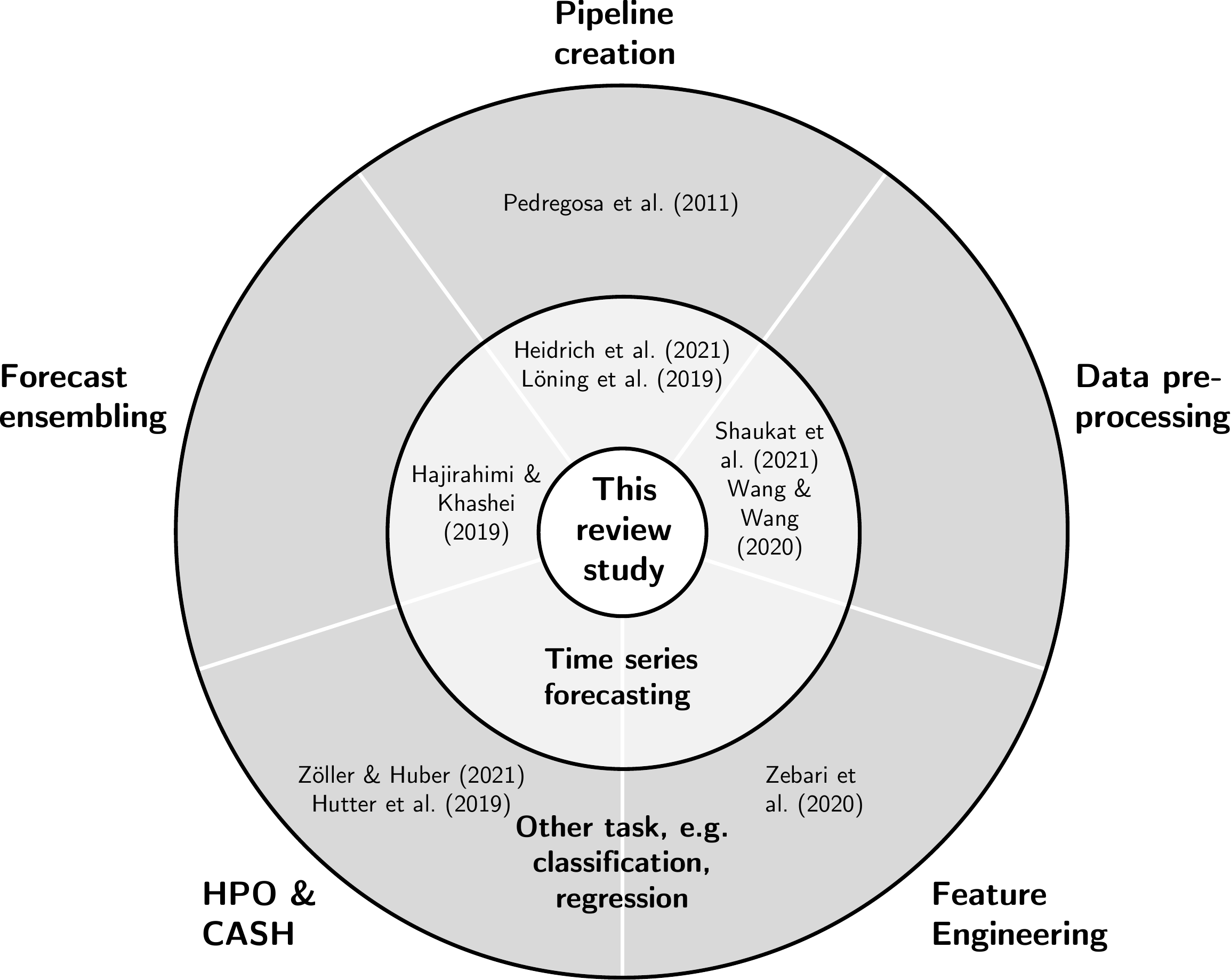}
	\scriptsize\\[3mm]
	HPO, HyperParameter Optimization;\\
	CASH, Combined Algorithm Selection and Hyperparameter optimization
	\caption{Related to this review are the fields of pipeline creation, data pre-processing, feature engineering, HyperParameter Optimization (HPO) \& Combined Algorithm Selection and Hyperparameter optimization (CASH), and forecast ensembling.}
	\label{fig:RelatedWork}
\end{figure}

In this context, the long-term objective towards full automation has motivated numerous researchers and led to promising research results in the fields related to this review study, shown in~\autoref{fig:RelatedWork}.
Several surveys and review studies analyze the automation of single forecasting pipeline sections such as pre-processing~\cite{Wang2020, Shaukat2021}, feature engineering~\cite{Zebari2020}, HPO and forecasting method selection~\cite{Hutter2019, Zoeller2021}, and forecast ensembling~\cite{Hajirahimi2019}.
Moreover, rather than focusing on the automated design of forecasting pipelines, existing studies on time series forecasting only consider the statistical or machine learning forecasting methods themselves~\cite{Gooijer2006, Han2019, SouhaibBenTaieb2012}.
However, a comprehensive review study on the entire automated time series forecasting pipeline that considers the families of both statistical and machine learning methods is lacking.

Therefore, the present paper analyzes the literature on automated time series forecasting pipelines to investigate how to automate the design process of forecasting models.
We consider literature from various research directions, including statistical forecasting, machine learning and deep learning.\footnote{
    In the following, we consider deep learning methods as part of the broader family of machine learning.
}
More specifically, we focus on the interaction and combination of automation methods within the pipeline sections considering both Automated Machine Learning (AutoML) and automated statistical forecasting methods.
For this purpose, we firstly present and systematically compare existing automation methods used in each pipeline section.
Secondly, we analyze the complete automated forecasting pipeline considering how many pipeline sections are automated in the literature and highlighting the interaction and dependencies between pipeline sections.
For both, we discuss the existing literature, identify potential problems, give recommendations, and suggest future research.

After describing the methodology in~\autoref{sec:Methodology} and a brief introduction to time series forecasting in~\autoref{sec:TimeSeriesForecasting}, the paper is organized by respective sections following the forecasting pipeline shown in~\autoref{fig:TimeSeriesPipeline} and concludes in~\autoref{sec:Conclusions}.

\newpage

%% file: content/03-methodology.tex
The methodology of this literature review applies the following fundamental steps suggested by Webster and Watson~\cite{Webster2002} for the identification of major contributions, their origin, and evolution.

\begin{description}
	\item[Literature Search:] A search-term-based exploration of research articles considering title, abstract, and keywords is conducted using \textit{Scopus}\footnote{
	    \href{https://www.scopus.com/}{https://www.scopus.com/}
	}, resulting in 359 hits.\footnote{\label{fn:SearchString}
	        The applied search strings and keywords are documented in the Supporting Information.}
		Potential predatory journals and publishers, and vanity press listed in \textit{Beall's List}\footnote{
		    \href{https://beallslist.net/}{https://beallslist.net/}
		} are excluded.
	\item[Literature Screening:] We screen the abstracts for relevance with the following criteria:
		i) the task type must be time series forecasting, and
		ii) at least one iterative element in the forecasting pipeline that previously required human intervention must be automated.
	\item[Backward-Forward Search:] We identify additional articles that cite or are cited by the articles passing the screening.
		The obtained candidates also undergo the screening procedure defined above.
		Backward-forward search yield a pool of 2152 additional papers, which the keyword filtering reduces to 242 articles.\footnoteref{fn:SearchString}
	\item[Review:] We present automation methods for each section of the forecasting pipeline and review the articles that pass the screening and the full text analysis.
	    After the abstract screening, 144 of the 601 papers remain, and 71 after analyzing the full text.
	    If there is similar work from a research group, we cite the article with the greatest methodological scope and articles that propose improvements, which reduces the included articles from 71 to 63 articles.
	\item[Discussion:] We discuss the contributions towards design automation individually for each section of the forecasting pipeline.
	    Afterward, we identify gaps and highlight future research directions by analyzing the coverage of the forecasting pipeline.
\end{description}

%% file: content/04-time-series-forecasting.tex
A time series $\left\{\mathbf{y}\left[k\right]; k = 1, 2 \ldots, K\right\}$ reflects a set of
$K \in \mathbb{N}^{>0}$ observations typically measured at equidistant points in time~\cite{Brockwell2016}.
A time series forecasting model $f(\cdot)$ estimates future values $\hat{y}$ for one or more time points -- the forecast horizon
	$H \in \mathbb{N}^{>0}$ -- using current and past values~\cite{GonzalezOrdiano2018}.
It is defined as
\begin{equation}
	\begin{split}
	\hat{y}[k+H] = f(	& y[k], \ldots, y\left[k-H_{1}\right], \\
						& \mathbf{u}^{T}[k], \ldots, \mathbf{u}^{T}\left[k-H_{1}\right], \\
						& \hat{\mathbf{u}}^{T}[k], \ldots, \hat{\mathbf{u}}^{T}\left[k-H_{1}\right], \\
						& \mathbf{w}); k, H, H_1 \in \mathbb{N}^{>0}; k > H_1,
	\end{split}
\end{equation}
where $H_1 \in \mathbb{N}^{>0}$ indicates the horizon for past values $k-H_1$, the vector $\mathbf{w}$ contains the model's parameters,
	the vector $\mathbf{u}^\top$ denotes values from exogenous time series, the vector $\hat{\mathbf{u}}^\top$ indicates that the exogenous values originate from another forecast,
	and $y$ represents values of the target time series~\cite{GonzalezOrdiano2018}.\footnote{
	The vectors that include past values can be sparse, i.\,e., only certain time points from $k$ to $H_1$ are included.
}
\bigbreak\medbreak
In time series forecasting, the following three families of methods exist: naïve methods, statistical methods, and machine learning methods.
The families and their main representatives are briefly introduced in the following.

\paragraph{Naïve Forecasting Methods}
The simplest method family of forecasting are naïve methods.
Common representatives are the averaging method, where the forecasts of all future values are equal to the average of historical data, and the Random Walk (RW) method, where the value of the last observation is used as forecast~\cite{Hyndman2021}.
For RW, modifications are available for data with drift (dRW) and seasonality (sRW).

\paragraph{Statistical Forecasting Methods}
More sophisticated than naïve methods are statistical methods that use statistics based on historical data to forecast future time series values.
The first representatives are AutoRegession (AR) methods.
The AutoRegressive Moving Average (ARMA) method~\cite{Box2016} assumes a linear relationship between the lagged inputs and is applicable if the time series is stationary.
If trends and seasonal characteristics are present, the time series is non-stationary, and the requirements for applying ARMA are not fulfilled.
To address this problem, the AutoRegressive Integrated Moving Average (ARIMA) method~\cite{Box2016} removes time series trends through differencing and the seasonal ARIMA (sARIMA) method eliminates the seasonality by seasonal differencing~\cite{Cheng2015}.

The second representatives are Exponential Smoothing (ES) methods, where the forecast is determined by a weighted average of past observations, with the weights decaying exponentially with their age~\cite{Hyndman2021}.
Simple Exponential Smoothing (SES) is a valid forecasting method for time series data without a trend or seasonal pattern.
For time series with a trend, SES is adapted to Double Exponential Smoothing (DES), and Triple Exponential Smoothing (TES) is suitable for time series with seasonality.
An extensive discussion of statistical forecasting methods can be found in reference~\cite{Gooijer2006}.
\paragraph{Machine Learning Forecasting Methods}
%
While most statistical forecasting methods are based on assumptions about the distribution of the time series data, machine learning methods have fewer restrictions in terms of linearity and stationarity~\cite{Cheng2015}.
In addition to statistical forecasting methods, which are specifically developed for time series forecasting, one can use regression methods based on machine learning to forecast multiple time points ahead using the following strategies, either solely or in combination~\cite{SouhaibBenTaieb2012}:
\begin{description}
	\item{\textbf{Recursive strategy}:}
	One trains a single regression model $f(\cdot)$ to forecast one time point ahead.
	In the operation, one recursively feedbacks the output value to the input for the next time point.
	\item{\textbf{Direct strategy}:}
	One trains multiple independent regression models $f_h(\cdot), h=1, \ldots H$, each to forecast the value at time $k + h$.
	The input is similar for each model.
	\item{\textbf{Multiple output strategy}:}
	One trains a single regression model to forecast the whole horizon $H$ at once.
	Consequently, the output is not a single value but a vector.
\end{description}

Representative machine learning methods are the Support Vector Regression (SVR), Decision Tree (DT)-based methods like the Gradient Boosting Machine (GBM), and Artificial Neural Networks (ANNs).
Reference~\cite{Han2019} gives an overview of machine learning and deep learning techniques applied to time series forecasting.

%% file: content/05-pre-processing.tex
Since most forecasting methods rely on assumptions about data properties, data pre-processing is of crucial importance.
Data pre-processing includes anomaly detection and handling, transforming the time series to make it stationary, and scaling the time series.
In the following sub-sections, automated methods for data pre-processing are introduced, and their utilization in forecasting pipelines is exemplified with the reviewed literature.

\subsection{Anomalies}
\label{ssec:Anomalies}
An anomaly is a value that significantly deviates from the rest of the time series~\cite{Chandola2009}.
Anomalies are induced by rare events or by errors in the data.
Apart from anomalous existing values, which we call outliers, anomalies also comprise missing values in the time series.
Both outliers and missing values can degrade the performance of forecasting methods or cause the training to fail.
Therefore, appropriate anomaly detection and handling are necessary.
\autoref{tab:anomaly} shows the summary of automated anomaly detection and handling methods used in the literature for time series forecasting pipelines.

\paragraph{Outlier Detection and Handling}
Automated outlier detection and handling aim to identify abnormal values and replace them with plausible values without human intervention.
Liu et al.~\cite{Liu2017} define an interval based on the global mean and the variance of the time series
\begin{equation}
    \left[\text{mean}(y)-\alpha_t\cdot\text{var}(y),\;\text{mean}(y)+\alpha_t\cdot\text{var}(y)\right],
    \label{eq:OutlierDetectionLiu}
\end{equation}
with the threshold $\alpha_t$ and the anomaly detection method considers values outside the interval as outliers.
Detected abnormal values are automatically substituted with the arithmetical averages of the nearest previous and posterior normal values.
However, anomalous values themselves bias the estimation of the mean and the variance, and the method is only valid for stationary time series.
Other authors tackle this weakness by calculating the local median instead of the global mean.
Martínez et al.~\cite{Martinez2019b}, Yan~\cite{Yan2012}, and Fan et al.~\cite{Fan2019} consider an observation as outlier if its absolute value is four times greater than the absolute medians of the three consecutive points before and after the observation.
However, only extreme values above the absolute medians are detected, extreme values below the absolute medians are not identified.
Widodo et al.~\cite{Widodo2016} apply the Hampel method, which automatically replaces any value that deviates from the median of its neighbors by more than three Median Absolute Deviations (MAD) with that median value.
Unlike previous methods, which use statistical measures, the anomaly detection and handling method of Maravall et al.~\cite{Maravall2015} is based on a forecasting model.
The method fits an ARIMA model, evaluates the MAD of the estimation residuals, and automatically replaces detected outliers with the forecast of the ARIMA model.
%
%
\begin{table}[tb]
	\centering
	\caption{Summary of automated anomaly detection and handling methods for data pre-processing in time series forecasting pipelines.}
	\begin{adjustbox}{max width=0.525\linewidth}
		\input{tables/anomaly}
	\end{adjustbox}
	\label{tab:anomaly}%
	\scriptsize\\[1mm]
	glob. global,
	imput. imputatio,
	loc. local,
	thresh. threshold,
	var. variance
\end{table}%
\paragraph{Missing Value Handling}
Automated missing value handling aims to reconstruct absent observations without human assistance.
The method of Fan et al.~\cite{Fan2019} automatically replaces missing values in the time series with the median of 12 consecutive points before and after the observation.
Yet, this method is prone to larger gaps of missing data.
The method of Züfle and Kounev~\cite{Zuefle2020} automatically imputes missing values by multiplying the known value one season before or after the missing value by the trend factor estimated between the day of the missing values and the day copied.
Using this procedure in chronological order allows the imputed values for the imputation of subsequent missing values.
After the missing value imputation, the authors apply a similar outlier detection method like \eqref{eq:OutlierDetectionLiu} with $\alpha_t=3$.
Unlike~\cite{Liu2017}, Züfle and Kounev~\cite{Zuefle2020} use the robust standard deviation between the \nth{1} and \nth{99} percentile of the data and replace the outliers by linearly interpolating between the two nearest non-anomalous values.\footnote{
    The authors disaggregated the time series into the seasonal, trend, and residual components and applied anomaly detection and handling on the stationary residual.
}

\subsection{Stationarity}
\label{ssec:Stationarity}
%
In a stationary time series $y[k]$, the statistical properties do not depend on the time of observation $k$, i.\,e., the distribution of $y[k,\ldots,k+s]$ is independent of $k$ for all $s$~\cite{Hyndman2021}.
Therefore, a time series with trends or seasonal patterns is not stationary because either the mean of the time series, its variance, or both change over time.
Since some statistical forecasting methods assume a stationary time series, their application to non-stationary time series requires an appropriate transformation.
Stationarity tests help to identify the type of non-stationarity and support the automatic selection of the appropriate transformation.
\autoref{tab:stationarity} shows the summary of automated stationarity testing and transformation methods used in the literature for time series forecasting pipelines, introduced in the following.
\begin{table}[tb]
	\centering
	\caption{Summary of automated stationarity testing and transformation methods for data pre-processing in time series forecasting pipelines.}
	\begin{adjustbox}{max width=0.6\linewidth}
		\input{tables/stationarity}
	\end{adjustbox}
	\label{tab:stationarity}%
	\scriptsize\\[1mm]
	s. seasonal,
	diff. differencing
\end{table}%
\paragraph{Autocorrelation and Differencing Transformations}
A first approach to automatically identify non-stationarities in time series is proposed by Tran and Reed~\cite{Tran2004} based on the AutoCorrelation Function (ACF) and the Partial AutoCorrelation Function (PACF).
The ACF and the PACF visualize the correlation of a time series with a delayed copy of itself.
The authors automatically detect decay patterns by calculating the average rate of change in the magnitude of high frequencies in the ACF and PACF, and consider rates of less than 10 percent as slow decay.
The slow decay patterns indicate trends in the time series.
To remove the trends, the time series is differenced by subtracting successive observations $d$ times.
After differencing, the ACF and PACF show significant peaks at regular intervals $s$, if the time series is seasonal.
To remove the seasonality, the time series is seasonally differenced $D$ times by subtracting observations separated by $s$.
Note that the ARIMA forecasting method explicitly include differencing as hyperparameter $d$ in the model structure and the sARIMA method additionally considers seasonal differencing with $D$ and $s$.

\paragraph{Frequency Filters}
Apart from the ACF and PACF, methods based on frequency filters are used to identify seasonality.
Bauer et al.~\cite{Bauer2020a} use the periodogram to automatically retrieve all frequencies within the time series, iterate over the found frequencies, and match each frequency with reasonable frequencies (e.\,g. daily, hourly, and yearly) with tolerance to determine seasonal frequencies.
Kourentzes and Crone~\cite{Kourentzes2010} propose the Iterative Neural Filter (INF) to automatically identify seasonal frequencies.
The filter distinguishes between stochastic and deterministic components and iteratively removes seasonalities, trends, and irregularities in the time series.
\newpage
\paragraph{Unit Root Tests}
Statistical unit root tests are used to identify non-stationarities before applying transformation methods.
The unit root tests used in the literature include
    the Kwiatkowski--Phillips--Schmidt--Shin (KPSS) test~\cite{KPSSTest} or the Cox-Stuart test~\cite{CoxStuartTest} for testing if the time series is stationary around a deterministic trend,
    the Augmented Dickey-Fuller (ADF) test~\cite{ADFTest} for the existence of stochastic trends in the time series,
    and the Canova-Hansen test~\cite{CHTest} or Osborn–Chui–Smith–Birchenhall (OCSN) test~\cite{OCSBTest} for the existence of seasonal patterns over time.
According to the trends and seasonalities detected in the unit root tests, the time series is automatically differenced or seasonally differenced, respectively.
\paragraph{Logarithm and Normality Transformations}
Apart from unit root testing, Maravall et al.~\cite{Maravall2015} and Liu et al.~\cite{Liu2017} propose log-level tests to automatically evaluate if a log-transformation of the time series is beneficial.
Amin et al.~\cite{Amin2012} use the Kolmogorov--Smirnov (KS) test~\cite{KSTest} to determine whether a time series is normally distributed -- if not, the log-transformation is applied.
Other authors apply the log-transformation without testing to reduce the variance~\cite{Anvari2016, Bandara2020} or apply various transformations until critical values are satisfied in t-tests of ACF and PACF~\cite{Lu2009}.
For achieving normality and stabilizing the variance, the Box-Cox transformation~\cite{BoxCoxTransformation} is often applied without preceding testing, like in references~\cite{Martinez2019b,Zuefle2020,Bauer2020a,Amin2012}.
\paragraph{Time Series Decomposition}
In addition to transformation operations to achieve stationarity, time series can be decomposed into the components trend, season, and irregular (i.\,e. the residual).
Afterward, each component can be handled by an individual forecasting model.
Prominent decomposition methods are the Seasonal and Trend decomposition using Loess (STL)~\cite{STLDecomposition} used in the references~\cite{Martinez2019b,Bauer2020a,Bandara2020}, and the additive or multiplicative decomposition applied by the authors of references~\cite{Widodo2016,Fildes2015}.

\subsection{Scaling}
\label{ssec:Scaling}
The scale of the time series influences the performance of many forecasting methods based on machine learning.
If the range of values is large, learning methods based on gradient descent may converge much slower or fail due to instability.
Additionally, in the case of multiple inputs with different scales, the inputs with larger variance dominate the others in the calculation of many distance measures~\cite{Sklearn}.

The general formalization for time series scaling is
\begin{equation}
    y' = \frac{y-a}{b}.
\end{equation}
\autoref{tab:scaling} shows the scaling methods used in the literature on automated forecasting pipelines compared to the family of the forecasting method.
A common approach is min-max scaling, where the time series are scaled in the range $[0,1]$ with $a=\min(y),\,b=\max(y)-\min(y)$.
While min-max scaling guarantees that all time series are scaled to the same range $[0,1]$, the Z-score normalization scales the time series to zero-mean and unit-variance with $a=\text{mean}(y),\,b=\text{var}(y)$.
Both scaling methods are sensitive to outliers and thus require a preceding anomaly detection and handling. Alternatively, robust scaling methods like \textit{scikit-learn's}  \textit{RobustScaler}\footnote{\href{https://scikit-learn.org/stable/modules/generated/sklearn.preprocessing.RobustScaler.html}{https://scikit-learn.org/stable/modules/generated/sklearn.preprocessing.RobustScaler.html}} can be used, which removes the median and scales the data according to the Inter-Quartile Range (IQR).
\begin{table}[h]
	\centering
	\caption{Summary of scaling methods for data pre-processing applied in time series forecasting pipelines.}
	\begin{adjustbox}{max width=0.6\linewidth}
		\input{tables/scaling}
	\end{adjustbox}
	\label{tab:scaling}%
\end{table}%
%
%
\subsection{Discussion}
\label{ssec:PreProcessingDiscussion}

We discuss data pre-processing as the first section of the automated forecasting pipeline, show possible problems, give recommendations, and suggest future research.

As shown in~\autoref{tab:stationarity}, automated stationarity testing and transformation methods are predominantly applied to pipelines using methods of the statistical forecasting family.
In contrast,~\autoref{tab:scaling} shows that time series scaling is mainly used in pipelines with methods of the machine learning family.
Both forecasting families require methods for automated anomaly detection and handling, shown in~\autoref{tab:anomaly}.

Regardless of their predominant use for automation, both transformation methods -- for achieving stationarity and scaling -- can be adversely affected by anomalies.
Therefore, it is essential that automated anomaly detection and handling is performed before time series transformations.
In automated outlier detection and handling, the main concern is that methods applied in the literature on automated forecasting pipelines focus on single outliers, although methods for consecutive outliers are widely available, e.\,g., \cite{BlazquezGarcia2021}.
Due to this focus, these methods may have a limited performance if several consecutive outliers or abnormal patterns are present.
For automated forecasting pipelines, we thus recommend the first step of pre-processing to be detecting and removing isolated outliers, followed by detecting consecutive anomalous values -- outliers or missing values -- and handling them with an appropriate method.
Ideally, the handling also considers domain-specific knowledge (e.\,g. \cite{Weber2021}).
\\
The second step of pre-processing, automated testing for stationarity and time series transformation, is only crucial if the pipeline uses a forecasting method that assumes a stationary time series.
For each condition -- deterministic and stochastic trends, seasonalities, and normality -- only one test must be applied because the use of multiple tests may lead to conflicting answers~\cite{Hyndman2021}.
For machine learning forecasting methods, however, these transformations are not mandatory.
Yet, a stationary time series may reduce the required complexity of the machine learning method.
\\
In the third step of pre-processing, we recommend using an appropriate time series scaling method to facilitate the training process of the respective forecasting method.

Given these recommendations, future work on automated forecasting pipelines should also consider domain knowledge to identify anomalies, e.\,g., if only positive values are valid or domain-adapted imputation to improve the reconstruction.
Additionally, it should be systematically evaluated whether stationarity transformations improve the forecast accuracy of machine learning methods if applied in the automated forecasting pipeline.

%% file: tables/anomaly.tex
\begin{tabular}{lp{8.7em}p{6em}p{5.95em}}
\toprule
\textbf{Ref.} & \textbf{Outlier\newline{}Detection} & \textbf{Outlier\newline{}Handling} & \textbf{Missing Value Handling} \\
\midrule
\cite{Liu2017} & glob. mean-var. thres. & average nearest &  \\
\cite{Martinez2019b, Yan2012} & loc. median thres. & average nearest &  \\
\cite{Fan2019} & loc. median thres. & average nearest & median imput. \\
\cite{Widodo2016} & loc. median-MAD\newline{} thres. & median nearest & \\
\cite{Maravall2015} & ARIMA-MAD thres. & ARIMA values & \\
\cite{Zuefle2020} & glob. median-robust-var. thres. & linear interpolation & copy-paste\newline{}imput. \\
\bottomrule
\end{tabular}%

%% file: tables/stationarity.tex
\begin{tabular}{llll}
\toprule
\textbf{Ref.} & \multicolumn{1}{p{8.21em}}{\textbf{Forecasting\newline{}Method(s)}} & \multicolumn{1}{p{9.2em}}{\textbf{Stationarity\newline{}Testing}} & \multicolumn{1}{p{7.625em}}{\textbf{Stationarity\newline{}Transformation}} \\
\midrule
\cite{Tran2004} & sARIMA & \multicolumn{1}{p{8.79em}}{ACF, PACF\newline{}pattern analysis} & diff., s. diff. \\
\cite{Bauer2020a} & Telescope & periodogram & Box-Cox, STL \\
\cite{Kourentzes2010} & MLP, TES & INF & INF \\
\cite{Crone2010} & MLP   & ADF, INF   & diff., INF \\
\cite{Alzyout2019} & ARIMA & KPSS  & diff. \\
\cite{Sekma2016a} & AR, VAR & KPSS & diff. \\
\cite{Hyndman2008} & autoARIMA & KPSS, Canova-Hansen & diff., s. diff. \\
\cite{Lu2009} & sARIMA & ACF, PACF t-test & log, diff., s. diff. \\
\cite{Maravall2015} & ARMA, sARIMA & \multicolumn{1}{p{9.2em}}{log-level, Kendall-Ord,\newline{}Pierce, Lytras, seasonal frequency peaks} & log, diff., s. diff. \\
\cite{Liu2017} & ARIMA & ADF, log-level & diff., log \\
\cite{Anvari2016} & sARIMA & \multicolumn{1}{p{9.2em}}{OSCB, KPSS, correlation, t-test, ADF} & diff., s. diff., log \\
\cite{Amin2012} & ARIMA, SETARMA & KS, KPSS & log, Box-Cox \\
\cite{Martinez2019b} & kNN   &       & diff., Box-Cox, STL \\
\cite{Fildes2015} & \multicolumn{1}{p{8.21em}}{autoARIMA, Theta, Damped, RW, sRW, SES, DES, TES} & Cox-Stuart & \multicolumn{1}{p{7.625em}}{multiplicative\newline{}decomposition} \\
\cite{Widodo2016} & \multicolumn{1}{p{8.21em}}{MKL} &       & \multicolumn{1}{p{7.625em}}{additive\newline{}decomposition} \\
\cite{Yan2012} & GRNN  & heuristic autocorrelation & diff., s. diff. \\
\cite{Bandara2020} & LSTM  &       & diff., log, STL \\
\bottomrule
\end{tabular}%

%% file: tables/scaling.tex
\begin{tabular}{p{6.835em}|lll}
\multicolumn{1}{l}{} & \multicolumn{1}{p{5em}}{\textbf{Min.-max.}} & \multicolumn{1}{p{5em}}{\textbf{Zero-mean}} & \multicolumn{1}{p{5em}}{\textbf{Z-score}} \\
\cmidrule{2-4}\multicolumn{1}{l|}{\textbf{Statistical}} & \cite{Sekma2016a} & \cite{Liu2017} & \cite{Dellino2018a} \\
\textbf{Machine\newline{}learning} & \multicolumn{1}{p{5.8em}}{\cite{Bauer2020b, Zuefle2019, Kourentzes2010, Crone2010, Martinez2019a, Violos2020, Maldonado2019, Yan2012, Panigrahi2020, Widodo2013, Sergio2016, Donate2013, Donate2014}} &       & \cite{Raetz2019, Ma2021, Widodo2013, Widodo2016, Sagaert2018} \\
\end{tabular}%

%% file: content/06-feature-engineering.tex
A time series feature reflects the observations of an explanatory variable in the process being forecast (i.\,e. the target variable).
\autoref{fig:TimeSeriesPipeline} shows feature engineering as a sub-section of the time series forecasting pipeline, consisting of extraction and selection of features.
\autoref{tab:feature-engineering} guides the following literature analysis.

\begin{table}[h]
	\centering
	\caption{Summary of feature engineering methods in automated time series forecasting pipelines.}
	\begin{adjustbox}{max width=1\textwidth}
		\input{tables/feature-engineering}
	\end{adjustbox}
	\label{tab:feature-engineering}%
	\scriptsize\\[0.5mm]
	cyc. cyc; tran. transformation, exo. exogenous, spec. spectral, frac. fractal, dim. dimension, var. variance, comb. combination
\end{table}%

\subsection{Feature Extraction}
\label{ssec:FeatureExtraction}

The feature space of the training data set contains for each explanatory variable a time series of the same resolution and length as the target variable.\footnote{
    If the features are originally in a different resolution or length, they must be transformed accordingly.
}
Feature extraction aims at automatically enriching the feature space with additional explanatory variables.
In the following, time series features are introduced, and their usage in automated forecasting pipelines is explored with literature references.

\paragraph{Lag Features}
In autocorrelated time series, past observations can be valuable explanatory variables to forecast the target variable.
Lag features provide values from prior time points for the forecasting method, i.\,e., at time point $k$, the model also processes values that date back a certain time horizon $H_1$.
Lag features are useful if the target variable has inertia that is significantly reflected in the resolution of the time series or if exogenous influences affect the target variable in periodic patterns~\cite{Raetz2019}.

\autoref{tab:feature-engineering} reveals that lag features are the most applied type of features.

\paragraph{Cyclic Features}
In the training data, the time stamps (e.\,g. \footnotesize\texttt{(YYYY-MM-DD hh:mm:ss)}\normalsize) of the target variable are unique and specify the sequence of the observations.
Cyclical patterns that humans can detect in this format, like the hour of the day, the day of the week, weekday and weekend, or month and year, cannot be processed by machine learning methods without encoding.
Cyclic features provide this cyclic relationship by ordinal, interval, or categorical encoding.
In the following, we exemplify these encodings for the day of the week (see~\autoref{tab:cyclic-encoding}).
Ordinal encoding assigns an integer numerical value to individual days.
For an interval encoding, one may utilize a periodical sine-cosine encoding to establish similarities between related observations,
    e.\,g., for encoding the day of the week, one adds two time series features that encode each weekday.\footnote{
        Two time series are necessary because, otherwise, the encoding would be ambiguous for several days.
}
A categorical encoding is achieved through so-called one-hot encoding.
In this example, we create a time series feature for each day of the week, being one on that day and zero otherwise.
Since the last category -- the Sunday -- is already implicitly represented if each other categorical feature is zero, one may omit an explicit feature.

\begin{table}[t]
	\centering
	\caption{Exemplification of cyclical encoding methods for the day of the week of a time series.}
	\begin{adjustbox}{max width=0.65\linewidth}
		\input{tables/cyclic-encoding}
	\end{adjustbox}
	\label{tab:cyclic-encoding}%
\end{table}%

In the literature reviewed, cyclic features are not widely applied.
Züfle and Kounev~\cite{Zuefle2020} apply Random Forest (RF) and GBM methods, using lag features and the cyclic features hour of the day, day of the week, and the exogenous feature holiday.
However, the encoding of the cyclic features is not described.
Sin-cos encoded cyclic features are used by Kourentzes and Crone~\cite{Kourentzes2010} as input for a MultiLayer Perceptron (MLP) forecasting method.

\paragraph{Transformation Features}
Time series transformations are widely used to make time series stationary, and can also be applied to extract explanatory variables that we term transformation features.
These features include calculating time derivatives, the decomposition into trend, seasonality and residual, moving averages~\cite{Cerqueira2021}, as well as applying mathematical operations on existing features, e.\,g., multiplication of exogenous features~\cite{Raetz2019}.

\paragraph{Exogenous Features}
In addition to features that are endogenously derived from the target variable, the forecast can be improved by using exogenous features if the target variable is subject to exogenous influences.
In addition, lag, cyclical, and transformation features can also be extracted from exogenous features.
Whether and which exogenous influences exist depends on the data domain.\footnote{
    We classify the data domain according to the following categories:
    economics \& finance, energy, nature \& demographics, human access, and other or unknown.
    The categories are adapted from the references~\cite{Makridakis2000, Bauer2021} and extended to cover the literature reviewed.
}

For example, energy data often depends on exogenous weather measures~\cite{Maldonado2019,Raetz2019,Son2015,Valente2020},
    human access data underlies the influences of public holidays and weather~\cite{Zuefle2020,Lowther2020},
    and sales data often correlates with economic indicators~\cite{Sagaert2018,Dellino2018b}.\footnote{
        Authors that apply forecasting methods on data from several domains mainly use univariate time series from forecasting competitions, where no exogenous time series are provided.
    }

\subsection{Feature Selection}
\label{ssec:FeatureSelection}

After automatically extracting several features, they typically undergo a selection to remove features that provide no additional information value, e.\,g., because of redundancy.
In the following, methods for automated feature selection are presented, and their application in forecasting pipelines is explored based on the reviewed literature.

\paragraph{Filter Methods}
Filter methods use metrics to rank single features or feature combinations and automatically select a promising feature set based on a threshold~\cite{Jovic2015}.
Hence, the filters rely on the general characteristics of the training data and are independent of the forecasting method and other subsequent sections in the forecasting pipeline.

The characteristics used for filtering are manifold.
Chakrabarti and Faloutsos~\cite{Chakrabarti2002} propose an automated filtering method to determine the optimal lag features based on a threshold of the time series' fractal dimension.
Martínez et al.~\cite{Martinez2019a} automatically select features by identifying significant lags in the PACF.
Kourentzes and Crone~\cite{Kourentzes2010} propose the INF method to identify seasonal frequencies in time series, which automatically selects lag and sin-cos encoded cyclic features.
The method of Bauer et al.~\cite{Bauer2020a} automatically filters frequencies of the time series using the periodogram with the threshold of a spectral value greater than $\SI{50}{\percent}$ of the most dominant frequency.
Additionally, cyclic features are extracted by calculating Fourier terms of the dominant frequencies.
The most dominant frequency is used to decompose the time series into trend, season, and residual components by STL.
Finally, the cyclic features and the seasonal component as a transformation feature are used to forecast the de-trended time series.

\paragraph{Wrapper Methods}
In contrast to filter methods, the wrapper methods assess candidate features based on the forecasting performance on a validation data set.
The best-performing feature set is selected by a search method, which tailors the feature set to the forecasting method or the entire forecasting pipeline, respectively.
The search methods and performance metrics used in the literature are diverse.
Independent from the chosen search method and metric, wrapper methods commonly take more computing time than filter methods, as training and validation require considerably more computing effort than calculating statistical measures.
However, empirical studies show that the computing effort pays off in a better performance of the wrapper approach compared to filter methods~\cite{Jovic2015}.
\newpage
In the literature reviewed, automated feature selection based on the performance of the forecasting pipeline is addressed by different methods.
The method of Yan~\cite{Yan2012} and Fan et al.~\cite{Fan2019} automatically determines the optimal lag features based on the forecast error by searching over candidate lags.
The candidate lags are either predefined individually (grid search) or drawn randomly between specified boundaries (random search).
Balkin and Ord~\cite{Balkin2000} and Martínez et al.~\cite{Martinez2019b} apply Forward Selection (FS) to automatically identify the optimal lag features.
Firstly, several forecasting pipelines are trained for each individual lag feature.
Secondly, the FS selects the best-performing lag feature and repeats the first procedure by adding another lag feature.
It retains the combination with the highest improvement and repeats the procedure until the forecasting performance stops increasing.
Besides search heuristics, optimization can be applied for feature selection.
For automatically selecting exogenous features, Lowther et al.~\cite{Lowther2020} adopt a Mixed Integer Quadratic Programming (MIQP) problem~\cite{Bertsimas2016} and Son and Kim~\cite{Son2015} apply Particle Swarm Optimization (PSO).
Three evolutionary search strategies for automatically selecting the optimal lag features of MLPs are evaluated by Donate et al.~\cite{Donate2013, Donate2014}, where the Estimation Distribution Algorithm (EDA) yields the best convergence speed and the lowest forecast error.

\paragraph{Embedded Methods}
Embedded methods integrate the feature selection into the training process of the forecasting method~\cite{Jovic2015}.\footnote{
    Since the embedded feature selection is specifically designed for the training algorithm of the respective forecasting method, the transfer to other forecasting methods is not straightforward.
}
During the training, the embedded feature selection commonly estimates the feature importance and weights the features accordingly.
Embedded methods require less computing effort than wrapper methods, but more than filter methods~\cite{Raetz2019}.

The embedded methods in the literature are specific to the forecasting method.
The approach of Panigrahi and Behera~\cite{Panigrahi2020} automatically determines the optimal lag features by a Differential Evolution Algorithm (DEA).
Instead of using gradient-based methods for training an MLP, the authors integrate the weight estimation into the DEA, aiming to increase the convergence speed.
Valente and Maldonado~\cite{Valente2020} consider the automated selection of lag and exogenous features by an FS embedded into the SVR training process.
The FS is based on a contribution metric that takes into account lags whose inclusion minimizes the metric.
An automated Backward Elimination (BE) for SVR using embedded kernel penalization is described by Maldonado et al.~\cite{Maldonado2019}.
Lag and exogenous features that are irrelevant for the forecasting performance are successively removed during training.

\paragraph{Hybrid Methods}
Hybrid methods aim to combine the advantages of the above methods.

Rätz et al.~\cite{Raetz2019} evaluate several filter, wrapper, embedded, and hybrid feature selection methods.
Based on their experiments, they propose to use a filter method to automatically remove all features with low variance in the first step.
Afterward, they apply Bayesian Optimization (BO) to assess the remaining feature candidate combinations, consisting of lag, transformation, and exogenous features based on the forecasting performance.
The approach of Widodo et al.~\cite{Widodo2016} filters significant lags using ACF, associates the remaining lag features with a kernel and automatically assigns appropriate weights to the kernels during training of the Multiple Kernel Learning (MKL) method.
Sagaert et al.~\cite{Sagaert2018} firstly filter candidate lag features using automated FS, similar to the stepwise search for the automated ARIMA design of Hyndman and Khandakar~\cite{Hyndman2008}.
Then, the identified set of lag features together with exogenous features are used as inputs of the embedded Least Absolute Shrinkage and Selection Operator (LASSO) method that automatically selects features by shrinking the coefficients of irrelevant ones.

\subsection{Feature Aggregation}
\label{ssec:FeatureAggregation}

Feature aggregation aims to transform the feature space into a low-dimensional representation while retaining the primary properties of the time series.
The transformation is advantageous in high-dimensional feature spaces to reduce processing time and avoid the curse of dimensionality.\footnote{
    As the dimensionality of the feature space increases, the available training data becomes sparse.
}
The Principal Component Analysis (PCA), e.\,g., maps the data to a lower-dimensional space, maximizing the variance in the lower-dimensional representation.\footnote{
    Since the PCA assumes a normal distribution, anomalies must be identified and handled beforehand and appropriate transformations need to be performed to obtain a stationary time series~\cite{Yang2005}.
}

Dellino et al.~\cite{Dellino2018b} pre-whiten the time series data with a PCA, i.\,e.,
    the PCA aggregates exogenous features and discards the principal components with a low variance that are assumed as noise.

\subsection{Discussion}
\label{ssec:FeatureEngineeringDiscussion}

We discuss feature engineering as the second section of the automated forecasting pipeline, highlight a potential issue, provide recommendations, and suggest future work.

As shown in~\autoref{tab:feature-engineering}, most automated pipelines that apply forecasting methods based on machine learning rely on lag features because -- unlike statistical forecasting methods -- they do not consider time lags implicitly.\footnote{
    A well-known exception is Recurrent Neural Networks (RNNs) that feedback input values within the neural structure to capture sequential relationships.
}
The sparse use of cyclic and transformation features can be explained because most of the analyzed forecasting pipelines already consider this information by automatically selecting appropriate lags.

In terms of automation, the majority of the literature combines the extraction of predefined features with an automated selection.
Regarding feature extraction, the primary concerns remain in the human-defined feature extraction since it requires experience and may be biased.
For this reason, extracting a large set of default features -- including lag, cyclical, and transformation features -- for the automated forecasting pipeline can be valuable since the subsequent automated feature selection removes irrelevant ones.
Additionally, exogenous features are a powerful opportunity to integrate domain-specific knowledge into the forecasts.\footnote{
    For multiple-point-ahead forecasts, future values of the exogenous time series are required if lag, cyclical, and transformation features are extracted from them; either through simultaneous forecasting or the integration of exogenous forecast results.
}
\\
In the automated feature selection, all four methods -- filter, wrapper, embedded, and hybrid -- are useful for removing irrelevant features.
While an embedded method is computationally beneficial by integrating the feature selection in the training process, it is specifically designed for a forecasting method.
Because of its independence from the forecasting method, we recommend combining a filter with a wrapper method (hybrid) to initially reduce the candidate features through filtering and then tailoring the feature selection to the forecasting method or pipeline, respectively.
\\
Since aggregated features often correlate with other features and the number of features can be reduced by an automated feature selection method, we do not consider automated feature aggregation necessary to reach a high forecast accuracy.
Moreover, the aggregation of features limits their interpretability regarding their information value for the forecast.

Based on these challenges, future work towards automated forecasting pipelines should consider the automated extraction of default endogenous features\footnote{
    Initial work on the automated extraction of endogenous features exists, e.\,g., \cite{Cerqueira2021,Barandas2020}.
}, extended by a domain-specific extraction to include exogenous features.

%% file: tables/feature-engineering.tex
\begin{tabular}{lllcccccccc}
\toprule
\multicolumn{1}{p{4.5em}}{\textbf{Ref.}} & \multicolumn{1}{p{6.335em}}{\textbf{Forecasting}} & \multicolumn{1}{p{4.54em}}{\textbf{Data}} & \multicolumn{4}{c}{\textbf{Feature Extraction}} & \multicolumn{3}{c}{\textbf{Feature Selection}} & \multicolumn{1}{p{4.625em}}{\textbf{Feature}} \\
      & \multicolumn{1}{p{6.335em}}{\textbf{Method(s)}} & \textbf{Domain} & lag & cyc. & tran. & exo. & filter & wrapper & embedded & \multicolumn{1}{p{4.625em}}{\textbf{Aggregation}} \\
\midrule
\cite{Cerqueira2021} & \multicolumn{1}{p{7em}}{AR, autoARIMA, ETS, TBATS} & several (5) & X     & -     & X     & -     &      &      &      &  \\
\cite{Zuefle2020} & GBM, RF & human access & X     & X     & -     & X     &       &      &      &  \\
\cite{Bauer2020a} & Telescope & several (5) &  -     & X     & X     & -     & spec. value &      &      &  \\
\cite{Chakrabarti2002} & kNN   & nature &   X     & -     & -     & -     & frac. dim. &      &      &  \\
\cite{Martinez2019a} & GRNN  & economics & X     & -     & -     & -     & PACF  &      &      &  \\
\cite{Yan2012} & GRNN  & energy & X     & -     & -     & -     &       & grid search &      &  \\
\cite{Fan2019} & ELM   & several (3) & X     & -     & -     & -     &       & random search &      &  \\
\cite{Balkin2000} & ANN, AR, RW & economics & X     & -     & -     & -     &       & FS    &      &  \\
\cite{Martinez2019b} & kNN   & economics & X     & -     & -     & -     &       & FS    &      &  \\
\cite{Lowther2020} & sARIMA & human access & -     & -     & -     & X     &       & MIQP  &      &  \\
\cite{Son2015} & SVR   & energy & -     & -     & -     & X     &       & PSO   &      &  \\
\cite{Donate2013, Donate2014} & MLP   & several (5) & X     & -     & -     & -     &       & EDA, DEA, GA &      &  \\
\cite{Panigrahi2020} & MLP   & several (5) & X     & -     & -     & -     &       &      & DEA   &  \\
\cite{Valente2020} & SVR   & energy & X     & -     & -     & X     &       &      & FS    &  \\
\cite{Maldonado2019} & SVR   & energy & X     & -     & -     & X     &        &      & BE    &  \\
\cite{Raetz2019} & \multicolumn{1}{p{6.335em}}{GBM, LASSO, MLP, SVR, RF} & energy & X     & -     & X     & X     &  low var. & BO    &      &  \\
\cite{Widodo2016} & MKL   & several (5) & X     & -     & -     & -     &  ACF   &      & kernel comb. &  \\
\cite{Kourentzes2010} & MLP   & human access & X     & X     & -     & -     &  INF   &      &      &  \\
\cite{Sagaert2018} & LASSO & economics & X     & -     & -     & X     &       & FS    & shrinkage &  \\
\cite{Dellino2018b} & sARIMA & economics & -     & -     & -     & X     &       &      &      & PCA \\
\bottomrule
\end{tabular}%

%% file: tables/cyclic-encoding.tex
\begin{tabular}{l|r|rr|rrrrrr|}
      & \multicolumn{1}{l|}{\textbf{Ordinal}} & \multicolumn{2}{c|}{\textbf{Sin-cos Interval}} & \multicolumn{6}{c|}{\textbf{One-hot Categorical}} \\
      & \multicolumn{1}{l|}{$x_\text{dow}$} & \multicolumn{1}{l}{$x_\text{sin}$} & \multicolumn{1}{l|}{$x_\text{cos}$} & \multicolumn{1}{l}{$x_\text{mon}$} & \multicolumn{1}{l}{$x_\text{tue}$} & \multicolumn{1}{l}{$x_\text{wed}$} & \multicolumn{1}{l}{$x_\text{thu}$} & \multicolumn{1}{l}{$x_\text{fri}$} & \multicolumn{1}{l|}{$x_\text{sat}$} \\
      \midrule
\textbf{Mon} & 0     & 1.000 & 0.000 & 1     & 0     & 0     & 0     & 0     & 0 \\
\textbf{Tue} & 1     & 0.623 & 0.782 & 0     & 1     & 0     & 0     & 0     & 0 \\
\textbf{Wed} & 2     & -0.223 & 0.975 & 0     & 0     & 1     & 0     & 0     & 0 \\
\textbf{Thu} & 3     & -0.901 & 0.434 & 0     & 0     & 0     & 1     & 0     & 0 \\
\textbf{Fri} & 4     & -0.901 & -0.434 & 0     & 0     & 0     & 0     & 1     & 0 \\
\textbf{Sat} & 5     & -0.223 & -0.975 & 0     & 0     & 0     & 0     & 0     & 1 \\
\textbf{Sun} & 6     & 0.623 & -0.782 & 0     & 0     & 0     & 0     & 0     & 0 \\
\end{tabular}%

%% file: content/07-hyperparameter-optimization.tex
Forecasting methods incorporate a wide range of hyperparameters about the forecasting model's structure, training regularization, and algorithm setup;
	parameters whose values are not directly derived from the data and must be selected by the data scientist.
The hyperparameter configuration $\mathbf{\lambda}$ includes all considered hyperparameters and their selected values~\cite{Feurer2019}.
By tailoring $\mathbf{\lambda}$ to the specific problem using HPO,
	one may improve the model performance over the default setting of common forecasting libraries~\cite{Raetz2019}.

\subsection{Performance Metrics and Validation Sample}
\label{ssec:ValidationDataMetric}

Most HPO methods assume that the performance of the model is quantifiable.
To evaluate the model performance in time series forecasting, measures from information theory and error measures are used.
Information Criteria (IC) measure the amount of information lost by a statistical model, taking into account the Goodness-Of-Fit (GOF) and the model complexity.
The less information a model loses, the higher the model performance. 
One IC is the Akaike Information Criterion (AIC)
\begin{equation}
    \text{AIC} = 2w - 2\ln\left(\hat{L}\right)
\end{equation}
with the number of estimated parameters $w$ and the model's maximum value of the likelihood function $\hat{L}$.
It rewards GOF but penalizes high numbers of model parameters.
The penalty is required since adding more parameters may increase the likelihood without being justified by the data (overfitting).
Another popular IC is the Bayesian Information Criterion (BIC)
\begin{equation}
    \text{BIC} = w\ln\left(K\right) - 2\ln\left(\hat{L}\right)
\end{equation}
that additionally considers the number of data points $K$.

Popular error measures are the Mean Squared Error (MSE)
\begin{equation}
    \text{MSE} = \frac{1}{N} \sum_{n=1}^{N}\left(y[n]-\hat{y}[n]\right)^{2},
\end{equation}
and the Mean Average Error (MAE)
\begin{equation}
    \text{MAE}=\frac{1}{N} \sum_{n=1}^{N}\left|y[n]-\hat{y}[n]\right|,
\end{equation}
where $\hat{y}_i$ is the forecast and $y_i$ the realized value.\footnote{
    There are further error measures that are derived from the MSE and the MAE, such as the Root MSE (RMSE), the Mean Absolute Scaled Error (MASE), the Mean Absolute Percentage Error (MAPE), and the symmetric MAPE (sMAPE).
    For their definitions, we refer to reference~\cite{Hyndman2021}.
}
The smaller the error of the model, the higher the model performance.
Error measures can be calculated \textit{in-sample} or \textit{out-of-sample}.
In-sample means that the forecast error is determined on data points that are part of the training data sample, while out-of-sample uses unseen points from the validation data sample.
To increase the robustness, cross-validation can be performed by splitting the data several times into different training and validation sub-sets and averaging the validation error across the folds.
The out-of-sample error validation is generally considered to be a more trustworthy empirical evidence, as in-sample error validation is prone to overfitting.\footnote{
    Since ICs try to prevent overfitting implicitly with the penalty term, they are computed in-sample.
}

\subsection{Optimization Methods}
\label{ssec:OptimizationMethods}

Given validation data and metrics, the hyperparameter configuration of forecasting methods can be optimized using different optimization methods.
\autoref{tab:hyperparameter-optimization} shows the summary of HPO methods for time series forecasting.
In the following sub-sections, automated HPO methods are introduced that are applied in the literature on forecasting pipelines.

\begin{table}[h!]
	\centering
	\caption{Summary of HyperParameter Optimization (HPO) methods in automated time series forecasting pipelines.}
	\begin{adjustbox}{max width=1\textwidth}
		\input{tables/hyperparameter-optimization}
	\end{adjustbox}
	\label{tab:hyperparameter-optimization}%
\end{table}%
\newpage
\subsubsection{Grid Search and Random Search}
\label{ssec:GridSearchRandomSearch}

\begin{figure}[!ht]
	\centering
	\begin{subfigure}[t]{0.28\linewidth}
		\includegraphics[width=0.90\linewidth]{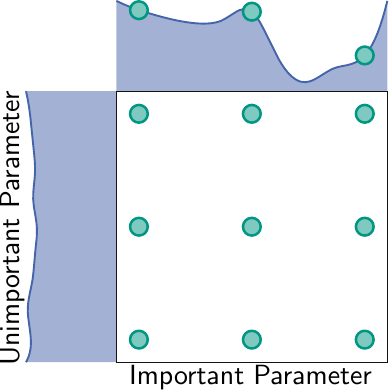}
		\caption{Grid search}
		\label{fig:GridSearch}
	\end{subfigure}
	\begin{subfigure}[t]{0.28\linewidth}
		\includegraphics[width=0.90\linewidth]{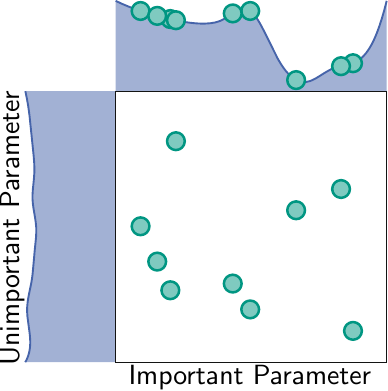}
		\caption{Random search}
		\label{fig:RandomSearch}
	\end{subfigure}
	\caption{
		Comparison of a \textit{grid search} and a \textit{random search} for minimizing a function with one important and one unimportant parameter~\cite{Feurer2019, Bergstra2012}.
	}
	\label{fig:GridSearchRandomSearch}
\end{figure}

The most elementary method for HPO is the exhaustive \textit{grid search}, where the data scientist defines a finite set of
	$\pi = 1,\ldots, \Pi$ hyperparameter values to be evaluated, resulting in a full factorial configuration space $\mathbf{\Lambda} \in \mathbb{R}^{\Pi}$.
As the grid search evaluates the Cartesian product of these sets,
	the number of computations $B$ grows exponentially with the dimensionality of $\mathbb{R}^{\Pi}$.
Hence, increasing the discretization resolution increases the computing effort substantially~\cite{Feurer2019}.
The \textit{random search}~\cite{Bergstra2012} is an alternative to the grid search.
It irregularly samples the hyperparameter set
	until a certain number of computations $B$ is exhausted.
The random search may performs better than the grid search if some hyperparameters are much more important than others.\footnote{
	The greater importance of a few hyperparameters over others applies in many cases, e.\,g.,~\cite{Raetz2019, Bergstra2012}.
}
\autoref{fig:GridSearchRandomSearch} shows a comparison of both methods with two hyperparameters and an equal number of computations $B$.
\bigbreak

Hyndman et al.~\cite{Hyndman2002} generalize the formulation of ES forecasting methods with the Error Trend Seasonality (ETS) method and suggest a grid search that automatically selects the hyperparameter configuration with the lowest in-sample AIC.
Utilizing grid search for HPO is also applied to statistical forecasting methods based on AR and Moving Averages (MAs) averages.
Sekma et al.~\cite{Sekma2016a} optimize the hyperparameters of an AR~$(p,s)$ and a Vector AutoRegression (VAR)~$(p,s)$ method, respectively, using grid search.
The hyperparameter configuration resulting in the minimal BIC, calculated in-sample, is automatically selected.
A similar HPO method for MA~$(q)$ methods is proposed by Svetunkov and Petropoulos~\cite{Svetunkov2018}, where the data scientist needs to define the in-sample optimization criterion.
An advanced method, combining pre-processing and HPO of ARIMA~$(p,d,q)$ methods is proposed by Alzyout et al.~\cite{Alzyout2019}.
Firstly, a stationarity test determines the differencing order $d$.
Secondly, the maximal AR-lag order $p_\text{max}$ and the maximal MA-lag order $q_\text{max}$ are determined by automatically evaluating the PACF and the ACF, respectively.
Finally, a grid search from $0$ to $p_\text{max}$ and $0$ to $q_\text{max}$ selects the optimal hyperparameter configuration based on in-sample AIC or BIC.
Combining pre-processing and HPO is also proposed by Hyndman and Khandakar~\cite{Hyndman2008}.
Firstly, the authors propose to automatically determine the differencing and seasonal differencing order $d$ and $D$ and the seasonal period $s$ with a stationarity test.
Secondly, instead of an exhaustive grid search over the hyperparameters $p,q,P$, and $Q$ of the sARIMA~$(p,d,q)(P,D,Q)_s$, they propose a two-stage grid search, reducing the number of evaluations.
In the first stage, four candidate hyperparameter configurations are evaluated, whose hyperparameters depend on the previously determined $s$.
The hyperparameter configuration with the smallest in-sample AIC value proceeds to the second stage, where $p,q,P$, and $Q$ are varied $\pm 1$.
If a configuration with a lower AIC value is found, it becomes the active configuration of stage two and the variation is repeated until the AIC stops improving.
Pedregal et al.~\cite{Pedregal2019} adopt this method except for the pre-processing step.
Instead of stationarity tests, the variance of the time series is minimized to identify the differencing orders $d$ and $D$.
%

Grid search is also applied for HPO of forecasting methods based on machine learning.
Sagaert et al.~\cite{Sagaert2018} determine the optimal shrinkage factor $\mathbf{\lambda}$ of the LASSO method for embedded feature selection in terms of the mean MAPE of a 10-fold cross-validation (out-of-sample).
Several authors apply a grid search to determine the optimal hyperparameter configuration of an SVR.
Son and Kim~\cite{Son2015} optimize the hyperparameters $C$ and $\gamma$ based on the RMSE,
    Maldonado et al.~\cite{Maldonado2019} and Valente and Maldonado~\cite{Valente2020} based on the MAPE.
The grid search used by Widodo et al.~\cite{Widodo2016,Widodo2013} is based on a 5-fold cross-validation, using the mean sMAPE value across folds as the forecast error measure.
They optimize the hyperparameters $C$ and $\varepsilon$ of the SVRs used in a MKL approach with embedded feature selection.
Combining automated feature selection and HPO is also suggested by Fan et al.~\cite{Fan2019}.
The authors identify the optimal lag features using random search and link the number of hidden neurons $N_\text{h}$ to the number of input neurons $N_\text{i}$.

\subsubsection{Bayesian Optimization}
\label{ssec:BayesianOptimization}

\begin{figure}[!ht]
	\centering
	\def\svgwidth{1\columnwidth}
	\includegraphics[width=0.6\linewidth]{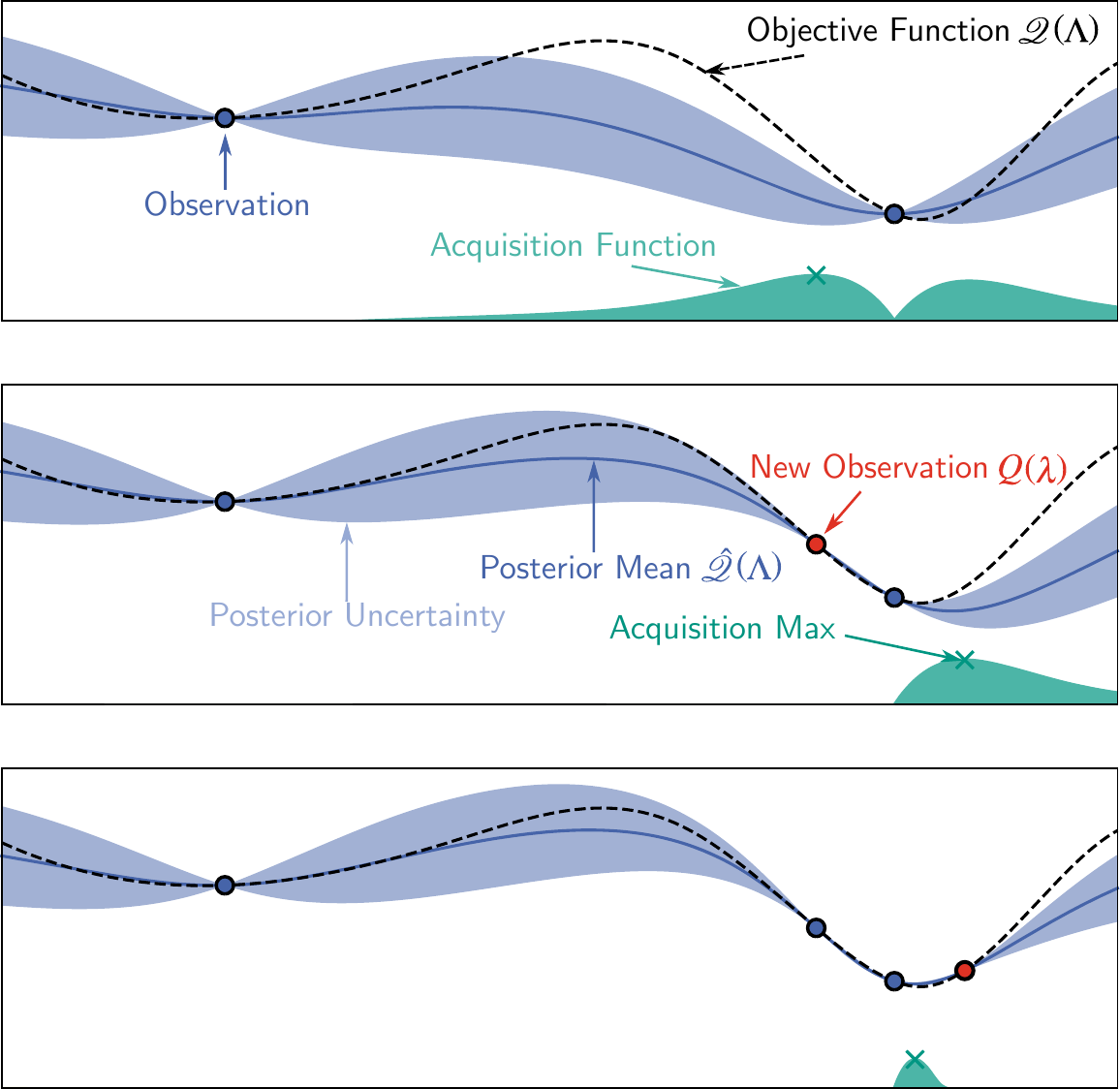}
	\caption{
		Two exemplary iterations of a Bayesian Optimization (BO) on a 1D function.
		The BO minimizes the predicted objective function $\hat{\mathcal{Q}}(\mathbf{\lambda})$ (blue line) by maximizing the acquisition function $\mathbb{E}\left[I(\mathbf{\lambda})\right]$ (green surface).
		The acquisition value is high where the value of $\hat{\mathcal{Q}}(\mathbf{\lambda})$ is low, and its predictive uncertainty (light blue interval) is high~\cite{Feurer2019}. The true objective function (dashed line) might lie outside of the predicted uncertainty interval.
	}
	\label{fig:BayesianOptimization}
\end{figure}

Rather than evaluating a finite search grid, BO explores and exploits the configuration space $\mathbf{\Lambda} = \Lambda_1 \times \Lambda_2, \ldots, \Lambda_H$ of $H$ hyperparameters $\Lambda$.
BO uses a probabilistic surrogate model to approximate the objective function $\mathcal{Q}$ that maps the forecasting pipeline's performance $Q$ on $\mathbf{\Lambda}$.
More specifically, an observation $Q(\mathbf{\lambda})$ of the objective function reflects the pipeline's performance with a particular hyperparameter configuration $\mathbf{\lambda} \in \mathbf{\Lambda}$ of the forecasting methods used.
In each iteration, the optimization updates the surrogate model with the new observation and uses an acquisition function to decide on the next hyperparameter configuration $\mathbf{\lambda} \in \mathbf{\Lambda}$ to be explored (see~\autoref{fig:BayesianOptimization}).
The acquisition function trades off the exploration against the exploitation of $\mathbf{\Lambda}$ by determining the expected benefit of different hyperparameter configurations using the probabilistic distribution of the surrogate model~\cite{Feurer2019}.

For surrogate modeling, various approaches exist, ranging from Gaussian Processes (GPs) and their modifications to machine learning approaches, e.\,g., RFs or Tree-structured Parzen Estimators (TPEs).
Feurer and Hutter~\cite{Feurer2019} recommend using a GP-based BO for configuration spaces with real-valued hyperparameters and computationally expensive training,
    and an RF or TPE-based BO for configuration spaces with categorical hyperparameters and conditions, e.\,g., the choice of a forecasting method and its conditional (sub-)configuration space.

Dellino et al.~\cite{Dellino2018a} apply a BO based on GP surrogate modeling to optimize the hyperparameters of the sARIMA~$(p,d,q)(P,D,Q)$ using an out-of-sample validation data set and the MAE.
They compare the BO to an exhaustive grid search, where the BO achieves lower forecast errors but requires more computing time.
In their experiment, however, the comparison is unequal as the configuration space of the BO is larger, and the best-performing model of the BO results in a hyperparameter configuration that the grid search does not evaluate.
\\
Bandara et al.~\cite{Bandara2020} use a BO with a GP surrogate model to optimize the recurrent neural architecture and training hyperparameters of a Long Short-Term Memory (LSTM).
The performance of each hyperparameter configuration is evaluated out-of-sample based on the MASE.
Rätz et al.~\cite{Raetz2019} optimize multiple forecasting methods, combining feature selection, HPO, and forecasting method selection using a BO based on a TPE surrogate model.
The hyperparameter configurations' performances are estimated using the mean MAE of a 5-fold out-of-sample cross-validation.

\subsubsection{Non-linear Programming}
\label{ssec:NonlinearProgramming}
Mathematical programming can be applied for HPO if calculating the performance metric is solvable in closed-form.
Non-Linear Programming (NLP) solves an optimization problem, where at least one of the objective functions or constraints is a non-linear function of the decision variables.
The objective function may be convex or non-convex, where non-convex NLP incorporates multiple feasible regions and multiple locally optimal solutions within them \cite{Bazaraa2006}.
Depending on the formulation of the objective function and its constraints, different solving methods are appropriate.

Bermúdez et al.~\cite{Bermudez2012} apply the generalized reduced gradient method to solve a multiobjective NLP.
They jointly minimize the in-sample RMSE, MAPE, and MAD to determine the smoothing parameters $\alpha, \beta, \gamma, \varphi$ and the number of periods of the seasonal cycle $p$ of the TES method.
Lowther et al.~\cite{Lowther2020} use MIQP to select suitable exogenous features for ARIMA.
They combine this selection with a grid search over the sparsity parameter $k$ of the MIQP formulation and the sARIMA hyperparameters $p,d,q,P,D,Q$.

\subsubsection{Heuristics}
\label{ssec:Heuristics}

A heuristic is an informed search technique that systematically explores a configuration space $\mathbf{\Lambda}$ subject to a constant search rule~\cite{Pearl1984}.

Tran and Reed~\cite{Tran2004} propose heuristics based on the ACF and PACF to determine the AR and MA lag order $p$ and $q$.
A similar approach is published by Amin et al.~\cite{Amin2012}.
To determine the transformation parameter $\theta$ of the Theta forecasting method, Spiliotis et al.~\cite{Spiliotis2020a} apply the Brent–Dekker method -- a root-finding method.
They determine the optimal $\theta$ for eight different trend $T$ and season $S$ configurations, and select the configuration that minimizes the in-sample MAE.
\\
Chakrabarti and Faloutsos~\cite{Chakrabarti2002} propose a heuristic to specify the number of nearest neighbors $k$ of the k-Nearest Neighbors (kNN) forecasting method after selecting the optimal lag features.
A training sample-related heuristic for determining the spread factor of the General Regression Neural Network (GRNN) is introduced by Yan~\cite{Yan2012}.

\subsubsection{Metaheuristics}
\label{ssec:MetaHeuristics}

Metaheuristics are strategies for guiding a search according to feedback from the objective function, previous decisions, and prior performance~\cite{Stuetzle1999},
    i.\,e., the searching behavior changes while exploring the configuration space $\mathbf{\Lambda}$.
Metaheuristics do not require assumptions about the objective function and can solve optimization problems where gradient-based methods fail.

\paragraph{Evolutionary Optimization}
\label{par:EvolutionaryOptimization}
Evolutionary Algorithms (EAs) comprise a wide range of population-based metaheuristics inspired by biological evolution~\cite{Baeck1996}.
A population of candidate hyperparameter configurations is evaluated using a fitness function to determine the performance of solutions.
Weak solutions drop out, while well-performing solutions evolve.
The mechanisms of selection and evolution differ between algorithms.

Genetic Algorithms (GAs) evolve a population of candidate hyperparameter configurations to explore and exploit the configuration space $\mathbf{\Lambda}$.
The hyperparameters of a candidate solution are encoded as genes in a chromosome.
In each generation, the fitness of the population is evaluated, and the chromosomes of individual candidates are modified to create a new generation -- the offspring.
The modification includes recombination and mutation and depends on an individual candidate's fitness.
A part of the population is retained and forms with the offspring the next generation.
DEAs differ from GAs in the mechanism of generating the offspring.
While in GAs an individual acts as parent to generate an offspring, the DEA adds the weighted difference between two chromosomes to create a new individual.
In this way, no separate probability distribution is required, making the algorithm self-organizing.
In EDAs, the population is replaced by a probability distribution over the choices available at each position in the chromosome of the individuals.
A new generation is obtained by sampling this distribution, avoiding premature convergence and making the representation of the population more compact.
\\
Donate et al.~\cite{Donate2013} evaluate a GA, a DEA, and an EDA for optimizing the hyperparameter configuration of an MLP, including the number of input and hidden neurons $N_\text{i}$ and $N_\text{h}$, as well as the training hyperparameters learning rate $\alpha$ and the weight initialization seed $s$.
The results of the experiments show that the DEA and the EDA require more than 100 generations to improve significantly over GA.
After 200 generations, the EDA achieves the lowest forecast error, followed by the DEA and the GA.
In a later publication~\cite{Donate2014}, the authors adapt the chromosome encoding and replace $s$ with the hyperparameter $\Delta_\text{max}$ of the used training algorithm.
In both publications, the fitness of each individual is evaluated by calculating the MSE on an out-of-sample validation data set.
Panigrahi and Behera~\cite{Panigrahi2020} apply a DEA to optimize $N_\text{i}$ and $N_\text{h}$ of an MLP, combining in-sample and out-of-sample validations.
The fitness of each individual (RMSE) is calculated in-sample, and the DEA is terminated when the RMSE on the validation data set increases\footnote{
    This regularization is also called \textit{early stopping}.
}, indicating the beginning of overfitting.

\paragraph{Particle Swarm Optimization}
\label{par:ParticleSwarmOptimization}

In PSO, a population of hyperparameter candidate configurations -- the swarm -- is evaluated.
The candidates move through the configuration space $\mathbf{\Lambda}$, where the movement of the swarm is guided by the best-performing candidates so far.

Sergio et al.~\cite{Sergio2016} apply PSO to optimize the hyperparameter configuration of multiple forecasting methods, combining the best hyperparameter configurations afterward to an ensemble.

\subsection{Diagnostic Checking}
\label{ssec:DiagnosticChecking}

Diagnostic checking evaluates the fitted forecasting model against criteria that indicate an adequate forecasting method configuration.
An adequate forecasting method configuration yields residuals $r[k] = y[k] - \hat{y}[k]$ with properties of white noise, i.\,e., the residuals are not autocorrelated having zero mean and finite variance~\cite{Hyndman2021}.
For statistical forecasting methods, the relationship between dependent and independent variables should be statistically significant.
That is, the parameters estimated in the fitting process describing this relationship are significantly different from zero.
If the forecasting method makes assumptions about the time series characteristics (e.\,g. stationarity), it is advisable to check whether the assumptions hold~\cite{Hyndman2021}.

Amin et al.~\cite{Amin2012} check the randomness of the residuals with a Box-Pierce test and the significance of the parameters of the ARIMA or Self Exciting Threshold AutoRegressive Moving Average (SETARMA) models with a t-test.
Furthermore, they test the invertibility and the stationarity, i.\,e., both the sum of the AR parameters and the sum of the MA parameters have to be smaller than one.
Analogously, Hwang et al.~\cite{Hwang2012} verify stationarity and invertibility.
For residual diagnostics, they additionally check the residual's correlations with the Ljung-Box test.
Similar to the above authors, Sekma et al.~\cite{Sekma2016a} test the parameters' significance and check if the residuals are white noise with the Ljung-Box test~\cite{LjungBoxTest}.

\subsection{Discussion}
\label{ssec:HPODiscussion}

We discuss HPO as the third section of the automated forecasting pipeline, identify a possible problem, give recommendations, and suggest future research.

For HPO, most automated forecasting pipelines apply grid search as shown in~\autoref{tab:hyperparameter-optimization}.
Directed search methods, e.\,g., evolutionary optimization and BO, are, however, used for HPO of forecasting methods with high computational training complexity, primarily including machine learning methods.
After an HPO, diagnostic checking is only applied sporadically and only for statistical forecasting methods.

With regard to automation, several authors link HPO and the preceding automated steps of pre-processing (\autoref{sec:PreProcessing}) and feature engineering (\autoref{sec:FeatureEngineering}).
One potential problem is the optimization of the differencing orders $d$ and $D$ of the ARIMA~$(p,d,q)$ and sARIMA~$(p,d,q)(P,D,Q)_s$ methods using IC metrics.
The differencing transformation affects the likelihood in IC metrics, making the metrics between different values of $d$ and $D$ not comparable~\cite{Hyndman2021}.
Therefore, $d$ and $D$ should be determined in the pre-processing section.
For the subsequent HPO, we assume a straightforward grid search to be sufficient since the computational training complexity of ARIMA and sARIMA methods is rather low, and the configuration space is small.
In contrast, for HPO of forecasting methods with high computational training complexity, e.\,g., ANNs, and large categorical and conditional configuration spaces, we recommend a BO based on TPE.

Given these recommendations, for automated forecasting pipelines in future work, we suggest that the automated analysis of residuals is also integrated into the pipeline using machine learning-based forecasting methods.

%% file: tables/hyperparameter-optimization.tex
\begin{tabular}{lllcccl}
\toprule
\multicolumn{1}{p{2.375em}}{\textbf{Ref.}} & \multicolumn{1}{p{9.6em}}{\textbf{Optimized Forecasting}} & \multicolumn{1}{p{7.8em}}{\textbf{Optimized Hyper-}} & \multicolumn{1}{c}{\textbf{Performance}} & \multicolumn{2}{c}{\textbf{Validation Sample}} & \multicolumn{1}{p{9.2em}}{\textbf{Optimization Method}} \\
      & \multicolumn{1}{p{9.875em}}{\textbf{Method(s)}} & \multicolumn{1}{p{12.21em}}{\textbf{parameters}} & \multicolumn{1}{c}{\textbf{Metric}} & in-sample & out-of-sample &       \\
\midrule
\cite{Tran2004} & sARIMA & $p,q$ &   & -     & -     &ACF \& PACF \\
\cite{Amin2012} & ARIMA, SETARMA & $p,q$ &   & -     & -     &\multicolumn{1}{p{8.5em}}{testing, ACF \& PACF, checking} \\
\cite{Sekma2016a} & AR, VAR & $p,s$ & BIC & X     & -     & \multicolumn{1}{p{8.5em}}{grid search, checking} \\
\cite{Svetunkov2018} & MA    & $q$   & user-defined & X     & -     & grid search \\
\cite{Alzyout2019} & ARIMA & $p,d,q$ & user-defined & X     & -     & \multicolumn{1}{p{8.5em}}{testing, ACF \& PACF, grid-search} \\
\cite{Hwang2012} & ARIMA & $p,d,q$ & GOF & X     & -     & grid search, checking \\
\cite{Hyndman2008} & sARIMA & $p,d,q,P,D,Q,s$ & AIC & X     & -     & \multicolumn{1}{p{8.5em}}{testing, two-stage\newline grid search} \\
\cite{Arlt2019} & sARIMA & $p,d,q,P,D,Q$ & user-defined & X     & X     & \multicolumn{1}{p{8.5em}}{testing, grid search,\newline checking} \\
\cite{Pedregal2019} & sARIMA & $p,d,q,P,D,Q,s$ & AIC & X     & -     & \multicolumn{1}{p{8.5em}}{var. minimization,\newline two-stage grid search} \\
\cite{Lowther2020} & sARIMA & $p,d,q,P,D,Q,s$ & user-defined & X     & X     & \multicolumn{1}{p{8.5em}}{two-stage MIQP \&\newline grid search} \\
\cite{Dellino2018a} & sARIMA & $p,d,q,P,D,Q$ & MAE & -     & X     & BO GP   \\
\cite{Hyndman2002} & ETS   & $E,T,S$ & AIC & X     & -     & grid search \\
\cite{Bermudez2012} & TES   & $\alpha,\beta,\gamma,\varphi,p$ &  RMSE, MAD & X     & -     & \multicolumn{1}{p{8.5em}}{multiobjective NLP} \\
 &    &  &  MAPE &      &      &  \\
\cite{Spiliotis2020a} & Theta & $\varphi$ & MAE & X     & -     & Brent \\
      &       & $T,S$ &       &       &  & \multicolumn{1}{p{8.5em}}{testing, grid search} \\
\cite{Villegas2019} & UC    & $T,S,I$ & BIC & X     &   -    & grid search \\
\cite{Chakrabarti2002} & kNN   & $k$   &  & -     & -     & heuristic \\
\cite{Sagaert2018} & LASSO & $\lambda$ & MAPE & -     & X     & grid search \\
\cite{Son2015} & SVR   & $C,\gamma$ & RMSE & -     & X     & grid search \\
\cite{Maldonado2019, Valente2020} & SVR   & $C,\gamma$ & MAPE & -     & X     & grid search \\
\cite{Widodo2013, Widodo2016} & MKL   & $C,\epsilon$ & sMAPE & -     & X     & grid search \\
      &       & $\text{kernel},\gamma$ &       &       &  & embedded \\
\cite{Yan2012} & GRNN  & $\text{spread}$ &  & -     & -     & heuristic \\
\cite{Panigrahi2020} & MLP   & $N_\text{i},N_\text{h}$ & RMSE & X     & X     & DEA    \\
\cite{Donate2014} & MLP   & $N_\text{i},N_\text{h},\alpha,\Delta_{\text{max}}$ & MSE & -     & X     & EDA   \\
\cite{Donate2013} & MLP   & $N_\text{i},N_\text{h},\alpha,s$ & MSE & -     & X     & GA, DEA, EDA \\
\cite{Fan2019} & ELM   & $N_\text{i},N_\text{h} $ & sMAPE & -     & X     & random search \\
\cite{Bandara2020} & LSTM  & $\text{epoch-size},\text{batch-size},$ & MASE & -     & X     & BO GP \\
      &       & $N_\text{h},\alpha,\text{epochs},\text{noise},\text{L2}$ &       &       &       &  \\
\cite{Sergio2016} & ANN, DBN, SVR & $\text{method-dependent}$ & MSE & -     & X     & PSO   \\
\cite{Raetz2019} & \multicolumn{1}{p{6.335em}}{GBM, LASSO, MLP, RF, SVR} & $\text{method-dependent}$ & MAE & -     & X     & BO TPE \\
\bottomrule
\end{tabular}%

%% file: content/08-selection-ensembling.tex
Not only optimizing hyperparameters of a forecasting method but also selecting the appropriate method is crucial for the forecast accuracy.
Consequently, the forecasting method selection is often combined with an individual HPO.\footnote{
    Selecting the optimal forecasting method and finding the optimal hyperparameter configuration is also called the Combined Algorithm Selection and Hyperparameter optimization (CASH) problem.
}
Forecast ensembling aims to bundle the forecasts of several methods, thereby reducing the impact of occasional poor forecasts -- which can even occur with the best-selected and optimally configured forecasting method.

\subsection{Forecasting Method Selection}
\label{ssec:ForecastingMethodSelection}

For automatically selecting the best-performing forecasting method, there are several approaches that we divide into heuristic, empirical, and decision model-based selection.
The selection is based on experience, a determined performance, or meta-features.
The determined performance reflects IC or error measures.
Meta-features describe properties of the time series to be forecast and provide meta-information, including statistical characteristics of the target time series, such as the skewness, kurtosis, and self-similarity; and domain information, such as physical properties of the system and environmental characteristics.
In the following, methods for automated forecasting method selection are presented, and their application in forecasting pipelines is examined based on the reviewed literature, guided by the summary in~\autoref{tab:selection-ensembling}.

\begin{table}[t]
	\centering
	\caption{Summary of method selection and ensembling methods in automated time series forecasting pipelines.}
	\begin{adjustbox}{max width=0.8\linewidth}
		\input{tables/selection-ensembling}
	\end{adjustbox}
	\label{tab:selection-ensembling}%
	\scriptsize\\[1mm]
	avg. averaging,
	dyn. dynamic,
	wgt. weighting
\end{table}%

\paragraph{Heuristic Forecasting Method Selection}

The heuristic selection of the forecasting method relies on fixed rules.
The basis of these rules is experience and statistical tests that examine the time series for certain characteristics.
Therefore, heuristic selection requires neither a determined performance nor meta-features.

Shcherbakov et al.~\cite{Shcherbakov2013} propose decision rules that consider the amount of available training data.
For small amounts of training data (i.\,e. less than 672 observations), a naïve method is selected that uses the previous day's value as the forecast.
For moderate amounts of training data (i.\,e. 672-2688 observations), a MA method is applied, whereas an ANN is selected for greater amounts of training data (i.\,e. more than 2688 observations).
Besides the amount of training data, the availability of calendar information and exogenous time series also determines the chosen forecasting method.

\paragraph{Empirical Forecasting Method Selection}

The empirical forecasting method selection determines the performance of several forecasting methods during training (in-sample) or on a validation data set (out-of-sample) and automatically selects the best-performing forecasting method.\footnote{
    For a detailed description of the in-sample and out-of-sample validation, refer to~\autoref{sec:HyperparameterOptimization}.
}

Balkin and Ord~\cite{Balkin2000} train an AR, an MLP, and a naïve RW method, and select the forecasting method with the smallest in-sample BIC.
Similarly, Sekma et al.~\cite{Sekma2016a} decide between AR and VAR forecasting methods based on the smallest in-sample BIC; Amin et al.~\cite{Amin2012} use the AIC to choose between ARIMA and SETARMA.
\\
An out-of-sample validation is used by Pereira et al.~\cite{Pereira2018}.
They calculate the MAE of six forecasting methods on a validation data set and select the method with the lowest MAE.
A similar selection strategy is used by Züfle and Kounev~\cite{Zuefle2020}. They select the best-performing candidate forecasting method in terms of the $R^2$ score on validation data.
Robustness can be increased by cross-validation.
Rätz et al.~\cite{Raetz2019} combine feature selection, HPO, and the selection of the optimal forecasting method using BO with the mean MAE over five folds.
\\
The effectiveness of in-sample and out-of-sample empirical forecasting method selection is evaluated by Fildes and Petropoulos~\cite{Fildes2015}.
They assess the following four empirical selection methods: i) minimal one-point-ahead in-sample MSE, ii) minimal one-point-ahead out-of-sample MAPE, iii) minimal $h$-point-ahead out-of-sample MAPE, and iv) minimal 1-18-points-ahead out-of-sample MAPE.
The latter method proves to be better than the 1-point-ahead validation and even than adjusting the validation to the corresponding forecast horizon $H$.

\paragraph{Decision Model-based Forecasting Method Selection}

Instead of a heuristic or empirical forecasting method selection, one may train a decision model to automatically select the optimal forecasting method.
As decision models, regression, classification, and clustering methods can be used to establish a relationship between performance information or meta-features and the optimal forecasting method.

In a decision model, the selection and aggregation of meta-features can improve decision accuracy.
The principle of feature selection and aggregation methods corresponds to the descriptions in~\autoref{sec:FeatureEngineering}.
Similar to the optimization of forecasts, the decision model can also be improved by HPO.
\autoref{tab:decision-model} gives an overview of meta-feature engineering and decision models applied in the literature.

\begin{table}[h]
	\centering
	\caption{Summary of meta-feature selection and aggregation methods for decision models to select forecasting methods.}
	\begin{adjustbox}{max width=0.73\textwidth}
		\input{tables/decision-model}
	\end{adjustbox}
	\label{tab:decision-model}%
	\scriptsize\\[1mm]
	class. classification,
	clust. clustering,
	reg. regression
\end{table}%

Taghiyeh et al.~\cite{Taghiyeh2020} propose a decision model-based selection method that relies on a classification method.
It selects the optimal forecasting method from a pool of candidates based on their in-sample and out-of-sample MSE.
None of the three classification methods, including Logistic Regression (LogR), DT, and Support Vector Machine (SVM), outperforms the others.
Kück et al.~\cite{Kueck2016} propose a decision model-based selection based on the out-of-sample sMAPE and meta-features.
They apply an MLP classifier as decision model and select its inputs using a grid search over 127 feature sets that include error measures and meta-features.

The approaches above have the disadvantage that all candidate forecasting methods must be trained on the target data set to determine the applied error measures.
Computing meta-features, in contrast, does not require training.
Hence, relying only on meta-features for decision model-based selection saves computing time.
Widodo and Budi~\cite{Widodo2013} propose a kNN classifier to select the forecasting method for a target time series based on the meta-features introduced in reference~\cite{Wang2009}.
In the design of the classifier, the authors apply FS for meta-feature selection.
Another approach based on the same meta-features is introduced by Scholz-Reiter et al.~\cite{ScholzReiter2014}.
They use a meta-feature aggregation instead of meta-feature selection, and a Linear Discriminant Analysis (LDA) as classification method to select the optimal forecasting method.
Shahoud et al.~\cite{Shahoud2020c} introduce statistical meta-features for different aggregation levels of the time series to extract characteristics at different time scales.
They select suitable meta-features by a BE and aggregate them using an autoencoder.
RF and ANN classifiers are compared for decision-making, both optimized with a grid search, where the ANN classifier achieves better performance in terms of selecting the best forecasting method.
In addition to statistical and time series meta-features, Cui et al.~\cite{Cui2016b} include domain information.
The domain-based meta-features describe physical properties of buildings for which energy consumption is to be forecast.
Bauer et al.~\cite{Bauer2020b} evaluate three types of decision models, i.\,e., classification, regression, and hybrid.
In the classification decision model, an RF is trained to map meta-features to the forecasting method with the lowest forecast error.
In the regression decision model, an RF learns how much worse each forecasting method is compared to the best method (i.\,e. forecast accuracy degradation).
The hybrid decision model combines this RF regression with an RF classifier that maps the RF regression prediction to the best method.
In the evaluation, the hybrid approach achieves the best performance in terms of forecast accuracy degradation.
Instead of training an individual model for each new time series, Bandara et al.~\cite{Bandara2020} suggest clustering time series using meta-features and training only one model for each cluster, which is applied to all time series in the cluster.

\subsection{Forecast Ensembling}
\label{ssec:ForecastEnsembling}

Forecast ensembling aims to improve the forecast robustness by bundling multiple forecasts of different models.
We differentiate forecast ensembling from ensemble learning methods that build an ensemble of weak models\footnote{
    The forecast of a weak model, e.\,g., a DT, is only slightly superior to a random estimate.
    Ensemble learning aims to combine many weak models to achieve a good estimate. 
}, such as RF and GBM.
Ensembling the forecasts from a pool of different forecasting models aims to avoid occasional poor forecasts, rather than outperforming the best individual forecasting model~\cite{Shaub2020}.
In the following, methods for ensembling in automated forecasting pipelines are introduced and exemplified using the reviewed literature, guided by the summary in~\autoref{tab:selection-ensembling}.

The benefit of forecast ensembling is empirically demonstrated in many cases.
For example, in the analysis of the so-called M3 forecasting competition~\cite{Makridakis2000}, averaging the model output of all submitted forecasting methods performs better than each individual method itself.
Simple forecast ensembling through averaging is used by Martínez et al.~\cite{Martinez2019b}.
They average the output of three kNN models with $k \in \left\{3,5,7\right\}$ after the identification of optimal lag features with FS.

To improve averaging, one may weight the forecasting methods according to the expected individual performance.
An ensemble can also be improved by only considering the k-best candidate methods, ranked by a forecasting method selection beforehand (i.\,e. heuristic, empirical, and decision model-based methods).

Selecting the candidate methods based on a heuristic forecasting method selection is proposed by Pawlikowski and Chorowska~\cite{Pawlikowski2020}.
They categorize the time series data of the M4 forecasting competition~\cite{Makridakis2020} in terms of their frequency, and the existence of a trend and seasonality.
Depending on the category, they select a distinct pool of candidate forecasting methods.
The hyperparameters of the candidate methods are optimized and the weight for each candidate is determined based on the sMAPE error, validated out-of-sample with a rolling origin evaluation.

The following authors use empirical forecasting method selection to consider only the k-best candidate methods in the ensemble.
Crone and Kourentzes~\cite{Crone2010} empirically determine the forecasting performance of candidate MLP architectures in a grid search ($N_\text{h}$, activation) with a rolling origin evaluation (out-of-sample) and average the outputs of the ten best candidates to reduce the impact of overfitting.
Kourentzes et al.~\cite{Kourentzes2019} propose an FS heuristic to decide on the number of ranked candidates to be considered for averaging.
They calculate the performance metric's rate of increase $C'$ assigned to each forecast and include all candidates until the first steep increase $C'>T$.
To detect the increase, they use the same approach used for detecting outliers in boxplots, i.\,e., $T = \text{Q3} + 1.5 \text{IQR}$, where $\text{Q3}$ is the \nth{3} quartile.

Instead of averaging the k-best candidates, Shetty and Shobha~\cite{Shetty2016} assign weights to the filtered candidates before averaging.
Candidates with low forecast errors $Q$ receive more weight, i.\,e.,
\begin{equation}
    w_{i}=\frac{\prod_{j=1}^{k} Q_{j}}{Q_{i} \sum_{j=1}^{k} Q_{j}}, \quad \sum_{i=1}^{k} w_{i}=1.
\end{equation}
Wu et al. \cite{Wu2020} propose a multi-objective optimization to determine the optimal weights for averaging candidates.
They apply the flower pollination metaheuristic to minimize
\begin{equation}
    \min\sum^4_{i=1}w_iQ_i, \quad \sum_{i=1}^{4} w_{i}=1
\end{equation}
with the error metrics $Q_i$ including the MAE, RMSE, and their relative formulations.

In the decision model-based method selection, one can also include weighted averaging directly in the decision model.
Montero-Manso et al.~\cite{MonteroManso2020} and Li et al.~\cite{Li2020} train a GBM classifier with softmax-transformed outputs corresponding to the weights of the candidates for averaging.
Similarly, Ma and Fildes~\cite{Ma2021} compare a Convolutional Neural Network (CNN) and a Fully Connected Neural Network (FCNN) classifier using output neurons with softmax activation to predict the weights for averaging.
Instead of softmax outputs, Züfle et al.~\cite{Zuefle2019} use an LR as a decision model.
Its output reflects the probability of how well the forecasting method fits the time series and determines the weight for averaging.
For filtering the k-best candidate methods, they post-process the LR output.

Above mentioned forecast ensembling methods with weighted average determine the weights statically, i.\,e., the weights do not change as the time series evolves.
Sergio et al.~\cite{Sergio2016} propose a dynamic weighting method for the ensembling of forecasts.
For every single forecast, the method searches for the k-nearest patterns in the training data similar to the given input data.
Three ensemble functions -- average, median, and softmax -- are evaluated on the found similar patterns, and the method with the best performance is chosen for the forecast.
A dynamic decision model-based approach is also introduced by Villegas et al.~\cite{Villegas2018}.
In their work, a binary SVM classifier is trained on performance metrics and meta-features to predict the best forecasting method from a pool of candidates for each forecast origin.
In their experiment, the dynamic selection achieves the best performance compared to the mean and the median ensembling of all candidate forecasting methods.

\subsection{Discussion}
\label{ssec:ForecastingMethodSelectionDiscussion}

We discuss the automated selection of the optimal forecasting methods as the fourth and forecast ensembling as the fifth section of the automated forecasting pipeline.
For both pipeline sections, we highlight potential problems, give recommendations, and suggest future work.

The automated selection includes heuristic, empirical, and decision model-based methods.
As shown in~\autoref{tab:selection-ensembling}, most empirical selection methods for the forecasting method are based on performance metrics, whereas decision models base their selection mainly on meta-features.
Considering automation, the selection of the forecasting method is often combined with an individual HPO of the candidate forecasting methods.
If multiple forecasting methods are evaluated in the selection, about half of the reviewed literature combines the selection with forecast ensembling.

In the automated selection, we notice different potential problems.
The heuristic forecasting method selection is based on straightforward decision rules.
To strengthen the evidence of the selection, however, we consider that performance metrics or meta-features are needed.
The empirical selection based on performance metrics requires a high computing effort as every candidate forecasting method needs to be fitted to the data.
The selection with a decision model based on meta-features reduces the computing effort but requires a sufficiently large and diverse data set for training the decision model.
Based on our analysis, a comprehensive benchmark comparing the computing effort and the forecast accuracy of heuristic, empirical, and decision model-based forecasting method selection is missing, and therefore no recommendation can be made.
Empirical evidence, however, exists for the forecast ensembling.
We recommend combining forecast ensembling with the forecasting method selection by determining the pool of candidate methods and ensemble weights.

Based on the challenges of automated forecasting method selection, future work could tailor decision models for specific domains using HPO.
Furthermore, the research could analyze if the selection of pre-trained forecasting methods is beneficial -- either for application to the new time series without adaptation or after re-training on the new data.
\newpage

%% file: tables/selection-ensembling.tex
\begin{tabular}{lp{15em}p{9.105em}cccc}
\toprule
\textbf{Ref.} & \multicolumn{1}{l}{\textbf{Forecasting}} & \multicolumn{1}{l}{\textbf{Forecasting Method}} & \multicolumn{2}{c}{\textbf{Selection Basis}}  & \multicolumn{2}{c}{\textbf{Forecast Ensembling}} \\
      & \multicolumn{1}{l}{\textbf{Method(s)}} & \multicolumn{1}{l}{\textbf{Selection}} & \multicolumn{1}{c}{pool} & \multicolumn{1}{c}{ensemble} \\
\midrule
\cite{Martinez2019b} & kNN   & \multicolumn{1}{l}{} & -     & -     & use all & avg. \\
\cite{Shcherbakov2013} & ANN, MA, sRW & heuristic & -     & -     &  &  \\
\cite{Pawlikowski2020} & autoARIMA, ETS, LR, RW, sRW, Theta & heuristic & -     & -     & use all & wgt. \\
\cite{Balkin2000} & ANN, AR, RW & empirical & X     & -     &  &  \\
\cite{Amin2012} & ARIMA, SETARMA & empirical & X     & -     &  &  \\
\cite{Sekma2016a} & AR, VAR & empirical & X     & -     &  &  \\
\cite{Pereira2018} & ARIMA DES, LR, dRW, SES, TES & empirical & X     & -     &  &  \\
\cite{Zuefle2020} & GBM, RF, Telescope & empirical & X     & -     &  &  \\
\cite{Raetz2019} & GBM, LASSO, MLP, RF, SVR & empirical & X     & -     &  &  \\
\cite{Fildes2015} & autoARIMA, Damped, RW, sRW, SES, DES, TES, Theta & empirical & X     & -     &  &  \\
\cite{Crone2010} & MLP   & empirical & X     & -     & k-best & avg. \\
\cite{Kourentzes2019} & ETS   & heuristic, empirical & X     & -     & k-best & avg. \\
\cite{Shetty2016} & autoARIMA, ETS, NNetAR, TBATS & empirical & X     & -     & k-best & wgt. \\
\cite{Wu2020} & ELM, ENN, FNN, GRNN, MLP, RBFNN & empirical & X     & -     & k-best & wgt. \\
\cite{Sergio2016} & DBN, MLP, SVR & heuristic, empirical & X     & -     & dyn. best & avg., wgt. \\
\cite{Taghiyeh2020} & autoARIMA, MA, SES, DES, TES, Theta & decision model & X     & -     &  &  \\
\cite{Kueck2016} & SES, DES, TES & decision model & X     & X     &  &  \\
\cite{Widodo2013} & autoARIMA, MKL, PR, SVR & decision model & -     & X     &  &  \\
\cite{ScholzReiter2014} & ANN, autoARIMA, ETS, LC, LL, RW & decision model & -     & X     &  &  \\
\cite{Shahoud2020c} & DT, GBM, LR, RF & decision model & -     & X     &  &  \\
\cite{Cui2016b} & ANN, GP, MARS, PR, RBFNN, SVR & decision model & -     & X     &  &  \\
\cite{Baddour2018} & autoARIMA, BSTS, ETS, GBM, Prophet, TBATS, Theta & decision model & -     & X     &  &  \\
\cite{Bauer2020b} & Telescope: Cubist, Evtree, GBM, NNetAR, RPaRT, SVR & decision model & -     & X     &  &  \\
\cite{Bandara2020} & LSTM  & decision model & -     & X     &  &  \\
\cite{MonteroManso2020} & AR, autoArima, ETS, NNetAR, RW, dRW, sRW, TBATS, Theta & decision model & -     & X     & use all & wgt. \\
\cite{Li2020} & autoARIMA, ETS, NNetAR, RW, dRW, sRW, STL-AR, TBATS, Theta & decision model & -     & X     & use all & wgt. \\
\cite{Ma2021} & autoARIMA, ELM, ETS, GBM, LR, RF, SVR & decision model & -     & X     & use all & wgt. \\
\cite{Zuefle2019} & autoARIMA, ETS, NNetAR, RW & decision model & -     & X     & k-best & wgt. \\
\cite{Villegas2018} & MA, IMA, White Noise & decision model & X     & X     &  &  \\
\bottomrule
\end{tabular}%

%% file: tables/decision-model.tex
\begin{tabular}{lp{10.25em}ccp{4.785em}p{3.43em}p{4.145em}}
\toprule
\textbf{Ref.} & \multicolumn{1}{l}{\textbf{Forecasting}} & \multicolumn{2}{c}{\textbf{Meta-Feature}} & \multicolumn{3}{c}{\textbf{Decision Model}} \\
      & \multicolumn{1}{l}{\textbf{Method(s)}} & \multicolumn{1}{c}{selection} & \multicolumn{1}{c}{aggregation} & \multicolumn{1}{c}{method} & \multicolumn{1}{c}{type} & \multicolumn{1}{c}{HPO} \\
\midrule
\cite{Taghiyeh2020} & autoARIMA, MA, SES, DES, TES, Theta &  &       & LogR, SVM, DT & class. &  \\
\cite{Kueck2016} & SES, DES, TES & grid search &       & MLP   & class. & grid search \\
\cite{Widodo2013} & autoARIMA, MKL, PR, SVR & FS    &       & kNN   & class. &  \\
\cite{ScholzReiter2014} & ANN, autoARIMA, ETS, LC, LL, RW &  & PCA & LDA   & class. &  \\
\cite{Shahoud2020c} & DT, GBM, LR, RF & BE    & autoencoder & MLP, RF & class. & grid search \\
\cite{Cui2016b} & ANN, GP, MARS, PR, RBFNN, SVR & PCA, Pearson & SVD & MLP   & class. & grid search \\
\cite{Baddour2018} & autoARIMA, BSTS, ETS, GBM, Prophet, TBATS, Theta & use all &       & RF    & class. & grid search \\
\cite{Bauer2020b} & Telescope: Cubist, Evtree, GBM, NNetAR, RPaRT, SVR & use all &       & RF    & reg., class. &  \\
\cite{Bandara2020} & LSTM  & use all &       & k-means, DBSCAN, Snob & clust. & embedded \\
\cite{MonteroManso2020} & AR, autoArima, ETS, NNetAR, RW, dRW, sRW, TBATS, Theta & use all &       & GBM   & class. & BO    \\
\cite{Li2020} & autoARIMA, ETS, NNetAR, RW, dRW, sRW, STL-AR, TBATS, Theta & learning &       & GBM   & class. & not described  \\
\cite{Ma2021} & autoARIMA, ELM, ETS, GBM, LR, RF, SVR & learning &       & CNN, FCNN & class. & grid search \\
\cite{Zuefle2019} & autoARIMA, ETS, NNetAR, RW & use all &       & LR    & reg.  \\
\cite{Villegas2018} & MA, IMA, White Noise & grid search &       & SVM   & class. & grid search \\
\bottomrule
\end{tabular}%

%% file: content/09-forecasting-pipeline.tex
The forecasting pipeline consists of the sections pre-processing, feature engineering, HPO, and forecasting method selection and ensembling (see~\autoref{fig:TimeSeriesPipeline}).

\subsection{Status Quo}
\autoref{tab:pipeline} shows that the analyzed literature covers different sections of the forecasting pipeline.\footnote{
    The scope and complexity with which the analyzed literature addresses the sections of the forecasting pipeline are analyzed in the previous Sections 4-7 of this paper.
}
Clustering the analyzed papers that cover the same sections of the forecasting pipeline yields 17 clusters.
Most of the 63 papers reviewed cover only two or three sections of the forecasting pipeline.
Only two papers cover four sections, and none of the papers reviewed covers all five sections of the forecasting pipeline.
\\
27 papers consider only statistical forecasting methods, 22 papers only machine learning methods, and 14 papers both.
A majority of the papers that include both families of forecasting methods consider forecasting method selection in their pipeline (12 of 14 papers).
\\
Of all the papers analyzed, most papers focus on pre-processing (35) and HPO (32), followed by the forecasting method selection (30).
Since statistical methods consider lag features implicitly in the model structure, only 21 papers explicitly deal with feature engineering.
The least attention is paid to the ensembling of forecasting methods, i.\,e., in only 11 papers.

\begin{table}[h]
	\centering
	\caption{Overview of the forecasting pipeline sections covered in the reviewed literature.
	}
	\begin{adjustbox}{max width=\textwidth}
		\input{tables/forecasting-pipeline}
	\end{adjustbox}
	\label{tab:pipeline}%
\end{table}%

\subsection{Discussion}
Based on the analysis of the status quo, we give the following recommendations to holistically automate the complete forecasting pipeline.
The optimization of the hyperparameters should be combined with the automated feature selection (clusters G, K, and P in \autoref{tab:pipeline}) because it is expected that each feature set candidate requires an individual hyperparameter configuration.
Moreover, combining an HPO and the automated selection of forecasting methods is advantageous as the best method is selected from a pool of candidates whose hyperparameters, in turn, are already optimized by an HPO (clusters I, L, P, and Q in \autoref{tab:pipeline}).
Also, the forecast ensembling should be combined with the automated selection of appropriate forecasting methods to eliminate poor candidates (clusters J, M, and Q in \autoref{tab:pipeline}).
Hereby, both statistical and machine learning-based forecasting methods should be considered, as the diversity of forecasting methods has the potential to increase the robustness of the results.

In addition to these recommendations, we discuss how automated forecasting pipelines are related to two other current research directions.
The first research direction is hybrid modeling, combining well-studied statistical methods with deep learning in time series forecasting.
More specifically, deep neural networks encode time-varying parameters for non-probabilistic forecasting methods or produce distribution parameters for probabilistic methods \cite{Lim2021}.
The second research direction is End-to-End (E2E) learning.
It aims to include all five sections of the forecasting model's design process into the training using gradient-based learning, e.g., embedded feature learning in a deep neural network \cite{Ma2021}.
While hybrid modeling only considers specific sections of the design process using deep learning, E2E learning uses one deep neural network to address several sections with appropriate layers.
As in automated forecasting pipelines, both hybrid modeling and E2E learning need to consider all five sections of the design process to obtain high-performing and robust forecasts.
Regarding automating the design process, emerging approaches such as neural architecture search and meta-learning \cite{Hutter2019} are promising for both research directions in the future.

Finally, we note that manual tailoring of forecasting models is still a common practice in time series forecasting competitions despite first successful applications of automated forecasting pipelines.
For example, in the most recent M5 forecasting competition \cite{Makridakis2021}, Liang and Lu \cite{Liang2020} provide evidence that their AutoML pipeline achieves competitive performance with moderate computing effort.
Therefore, we assume that the potential of the automated design is not fully utilized yet in time series competitions.

%% file: tables/forecasting-pipeline.tex
\begin{tabular}{lccccccc}
\toprule
\multicolumn{1}{p{2.355em}}{\textbf{Ref.}} & \multicolumn{1}{p{3.715em}}{\textbf{Pre-}} & \multicolumn{1}{p{4.32em}}{\textbf{Feature}} & \multicolumn{1}{p{6.035em}}{\textbf{Hyperparameter}} & \multicolumn{1}{p{7.25em}}{\textbf{Forecasting}} & \multicolumn{1}{p{4.18em}}{\textbf{Forecast}} & \multicolumn{1}{p{3.035em}}{\textbf{Covered}} & \multicolumn{1}{p{2.645em}}{\textbf{Cluster}} \\
      & \multicolumn{1}{p{3.715em}}{\textbf{processing}} & \multicolumn{1}{p{4.32em}}{\textbf{Engineering}} & \multicolumn{1}{p{6.035em}}{\textbf{Optimization}} & \textbf{Method Selection} & \textbf{Ensembling} & \multicolumn{1}{p{3.035em}}{\textbf{Sections}} &       \\
\midrule
\cite{Liu2017,Lu2009,Anvari2016,Cerqueira2021} & X     & -     & -     & -     & - & $\bullet$ & A     \\
\cite{Hyndman2002,Svetunkov2018,Pedregal2019,Bermudez2012,Hwang2012,Villegas2019,Arlt2019} & -     & -     & X     & -     & - & $\bullet$     & B     \\
\cite{Pereira2018,Shcherbakov2013, Shahoud2020c,Taghiyeh2020,Villegas2018,Kueck2016,Cui2016b,ScholzReiter2014,Baddour2018} & -     & -     & -     & X     & - & $\bullet$    & C     \\
\cite{Kourentzes2010,Martinez2019a,Bauer2020a} & X     & X     & -     & -     & - & $\bullet$ $\bullet$    & D     \\
\cite{Hyndman2008,Maravall2015,Dellino2018a,Alzyout2019,Tran2004} & X     & -     & X     & -     & - & $\bullet$ $\bullet$    & E     \\
\cite{Fildes2015,Violos2020} & X     & -     & -     & X     & - & $\bullet$ $\bullet$    & F     \\
\cite{Lowther2020,Chakrabarti2002,Valente2020,Son2015} & -     & X     & X     & -     & - & $\bullet$ $\bullet$    & G     \\
\cite{Balkin2000} & -     & X     & -     & X     & - & $\bullet$ $\bullet$    & H     \\
\cite{Spiliotis2020a,Bandara2020} & -     & -     & X     & X     & - & $\bullet$ $\bullet$    & I     \\
\cite{Li2020,MonteroManso2020,Shetty2016,Kourentzes2019,Wu2020,Pawlikowski2020} & -     & -     & -     & X     & X & $\bullet$ $\bullet$   & J     \\
\cite{Maldonado2019,Yan2012,Panigrahi2020,Sagaert2018,Widodo2016,Donate2014,Donate2013,Fan2019,Dellino2018b} & X     & X     & X     & -     & - & $\bullet$ $\bullet$ $\bullet$   & K     \\
\cite{Amin2012,Sekma2016a,Widodo2013} & X     & -     & X     & X     & - & $\bullet$ $\bullet$ $\bullet$   & L     \\
\cite{Ma2021,Crone2010,Zuefle2019} & X     & -     & -     & X     & X & $\bullet$ $\bullet$ $\bullet$   & M     \\
\cite{Martinez2019b} & X     & X     & -     & -     & X & $\bullet$ $\bullet$ $\bullet$   & N     \\
\cite{Zuefle2020,Bauer2020b} & X     & X     & -     & X     & - & $\bullet$ $\bullet$ $\bullet$   & O     \\
\cite{Raetz2019} & X     & X     & X     & X     & - & $\bullet$ $\bullet$ $\bullet$ $\bullet$    & P     \\
\cite{Sergio2016} & X     & -     & X     & X     & X & $\bullet$ $\bullet$ $\bullet$ $\bullet$    & Q     \\
\bottomrule
\end{tabular}%

%% file: content/10-conclusions.tex
Time series are collected in many domains, and forecasting their progression over a certain future period is becoming increasingly important for many use cases.
This rapidly growing demand requires making the design process of time series forecasts more efficient by automation;
that is, automating design decisions within each section of the forecasting pipeline or automatically combining methods across pipeline sections.
Although the first aspect of design automation has already been considered by various researchers, understanding how various automation methods interact within the pipeline and how they can be combined is a critical open question.
%
Therefore, the present paper considers the automation of each of the five sections in the time series forecasting pipeline.
It also investigates the corresponding literature in terms of the interaction and combination of automation methods within the five pipeline sections, incorporating both Automated Machine Learning (AutoML) and automated statistical forecasting methods.

Besides various specific insights related to each pipeline section that we discuss throughout the corresponding sections, we find on a general level that the majority of the 63 papers only cover two or three of the five forecasting pipeline sections.
Therefore, we conclude that there is a research gap regarding approaches that holistically consider the automation of the forecasting pipeline, enabling the large-scale application to use cases without manual and time-consuming tailoring.

Besides the holistic automation, future work should research the adaption of the presented automation methods for probabilistic time series forecasts.
Concurrently, the automated forecasting pipelines should be validated and tailored to particular use cases as an initial starting point before generalizing them to universal automated forecasting pipelines.
Furthermore, given the insights on the combination and interaction of automation methods in the pipeline, future work should examine their performance (e.\,g. in forecasting competitions).
This performance evaluation would benefit if future work on automated forecasting pipelines would be open source -- both the implementation of the evaluated methods and the data sets used for the evaluation.
Moreover, open-source publishing should promote the adoption of automated pipelines for time series forecasting, thus leveraging the great potential of improving the design efficiency through automation, achieving a high forecasting performance and a robust operation.\\
Overall, the present paper focuses on the automated design of forecasting pipelines.
Therefore, to fully automate the entire forecasting process, future work should also consider the automated application of the resulting forecasting models, including performance monitoring during operation and model adaption.\footnote{
    For the fusion of automated design and automated application, automation levels are defined by Meisenbacher et al. \cite{Meisenbacher2021}.
}